\documentclass[twoside]{article}

\usepackage[accepted]{aistats2023}
\usepackage[round]{natbib}

\usepackage[utf8]{inputenc} % allow utf-8 input
\usepackage[T1]{fontenc}    % use 8-bit T1 fonts
\usepackage{hyperref}       % hyperlinks
\usepackage{url}            % simple URL typesetting
\usepackage{booktabs}       % professional-quality tables
\usepackage{amsfonts}       % blackboard math symbols
\usepackage{nicefrac}       % compact symbols for 1/2, etc.
\usepackage{microtype}% microtypography
\usepackage[inline]{enumitem}
\usepackage{xcolor}         % colors
\usepackage{tikz}
\usetikzlibrary{bayesnet}
\usepackage{wrapfig}
\usepackage{multirow}
\usepackage{multicol}
\usepackage{caption}
\usepackage{subcaption}
\usepackage{xspace}

\usetikzlibrary{fadings}
\usetikzlibrary{patterns}
\usetikzlibrary{shadows.blur}
\usetikzlibrary{shapes}

\newcommand{\ours}[0]{BaCaDI\xspace} % \xspace makes spacing correct after the custom command and no need to add "~"
\usepackage{amsmath,amsfonts,bm}

\def\eqref#1{equation~\ref{#1}}
\def\1{\bm{1}}
\newcommand{\train}{\mathcal{D}}

\newtheorem{proposition}{Proposition}

\def\rvx{{\mathbf{x}}}

\def\evx{{x}}

\DeclareMathAlphabet{\mathsfit}{\encodingdefault}{\sfdefault}{m}{sl}
\SetMathAlphabet{\mathsfit}{bold}{\encodingdefault}{\sfdefault}{bx}{n}

\newcommand{\E}{\mathbb{E}}

\newcommand{\R}{\mathbb{R}}

\newcommand{\sigmoid}{\sigma}

\DeclareMathOperator*{\argmin}{arg\,min}

\usepackage{algorithm}
\usepackage[noend]{algpseudocode}

\newcommand\erdosrenyi[0]{\text{Erd{\H{o}}s-R{\'e}nyi}}

\newcommand{\G}[0]{{\bf{G}}}
\newcommand{\Z}[0]{{\bf{Z}}}
\newcommand{\pa}[1]{{\text{pa}_{\G}(#1)}}

\newcommand{\param}[0]{{\bf{\Theta}}}
\newcommand{\intvGam}[0]{{\bf{\Gamma}}}
\newcommand{\I}[0]{\mathcal{I}}
\newcommand{\D}[0]{\mathcal{D}}
\newcommand{\Itar}[0]{I^{\text{tar}}}
\newcommand{\phib}[0]{\boldsymbol{\phi}}

\newcommand{\vspaceparagraph}{\vspace{0pt}}
\newcommand{\vspacecaption}{\vspace{-6pt}}
\newcommand{\vspacecaptionlow}{\vspace{0pt}}
\newcommand{\vspacesubcaption}{\vspace{-3pt}}
\newcommand{\vspacesubcaptionlow}{\vspace{0pt}}
\newcommand{\vspacefigabove}{\vspace{0pt}}
\newcommand{\vspacefigcaption}{\vspace{-18pt}}
\newcommand{\vspacefigcaptionmid}{\vspace{0pt}}
\newcommand{\vspacefigcaptionlow}{\vspace{0pt}}
\newcommand{\vspaceequation}{\vspace{0pt}}

\allowdisplaybreaks
\allowdisplaybreaks
\begin{document}

\runningtitle{BaCaDI: Bayesian Causal Discovery with Unknown Interventions}

\runningauthor{Alexander H\"agele, Jonas Rothfuss, Lars Lorch, Vignesh Ram Somnath, Bernhard Sch\"olkopf, Andreas Krause}

\twocolumn[

\aistatstitle{BaCaDI: Bayesian Causal Discovery with Unknown Interventions}

\aistatsauthor{ Alexander H\"agele$^{*,1}$ \And Jonas Rothfuss$^{1}$\And Lars Lorch$^{1}$\And Vignesh Ram Somnath$^{1,3}$\AND
    Bernhard Sch\"olkopf$^{2,1}$\And Andreas Krause$^{1}$}
\vspace{10pt}

\aistatsaddress{$^{1}$Department of Computer Science, ETH Z\"urich, Switzerland\\
  $^{2}$Max Planck Institute for Intelligent Systems, T\"ubingen, Germany\\
  $^{3}$IBM Research Z\"urich, Switzerland\\
  *Correspondence to: \texttt{ahaegele@ethz.ch}\\}]

\begin{abstract}
%\looseness -1 
Inferring causal structures from experimentation
is a central task in many domains. For example, in biology, recent advances allow us to obtain single-cell expression data under multiple interventions such as drugs or gene knockouts. However,  the targets of the interventions are often uncertain or unknown and the number of observations limited. As a result, standard causal discovery methods can no longer be reliably used.
To fill this gap, we propose a Bayesian framework (\ours) for discovering and reasoning about the causal structure that underlies data generated under various unknown experimental or interventional conditions.
\ours is fully differentiable, which 
% operates in the continuous space of latent probabilistic representations of both causal structures and interventions.
% This 
allows us to %consistently 
infer the complex joint posterior
over the intervention targets and the causal structure
via efficient gradient-based variational inference.
% and to reason about the {epistemic uncertainty} in the predicted structure.
In experiments on synthetic causal discovery tasks and simulated gene-expression data, \ours outperforms related methods in identifying causal structures and intervention targets. 
% Finally, we demonstrate that
% %, thanks to its rigorous Bayesian approach, 
% our method provides well-calibrated uncertainty estimates.
\end{abstract}
\vspace{-4pt}

\vspacecaption
\section{INTRODUCTION}
\vspacecaptionlow
\looseness -1 Identifying causal dependencies via empirical observation and experimentation is a problem of fundamental scientific interest. If we understand the causal mechanisms that govern a system of interest, we can predict its behavior when parts of system are actively manipulated from outside. 
For instance, if we understand the causal structure of a gene regulatory network, we can predict the effect of drugs or gene knockouts more reliably \citep{meinshausen2016methods}.
While identifying the true causal structure from observational data is often impossible \citep{pearl_2009, Peters2017}, intervening on variables in the system and observing the outcome can fully identify the true causal structure in the large sample limit and the absence of latent confounders \citep{eberhardt2005number,eberhardt2006n,hyttinen2013experiment}.

Intervening in real-world systems, however, is often costly and imprecise, resulting in a limited number of observations and uncertain or unknown intervention targets and effects.
For example, recent advances in biology enable the collection of single-cell gene expression data under various interventions such as chemical compounds or gene knockouts \citep{srivatsan2020massively, mcfarland2020multiplexed}.
Given this significant epistemic uncertainty, causal structure learning becomes a complex joint inference problem over the structure and the unknown intervention targets that ultimately allow its identification. 
However, existing causal discovery methods that leverage interventional data typically assume full knowledge of the targets or the statistical effect of the interventions \citep{hauser2012characterization, wang2017permutation, yang2018characterizing}.
Conversely, recent methods that are able to deal with imperfect or unknown interventions do not account for epistemic uncertainty during inference \citep{mooij2016multiple,ke2019dsdi,brouillard2020differentiable,squires2020permutationbased}. Thus, they frequently fail in realistic scenarios where interventional data is scarce.

Addressing these shortcomings, we introduce {\em Bayesian Causal Discovery with unknown Interventions (\ours)}, a fully Bayesian approach for inferring the complete joint posterior over the causal structure, the parameters of the causal mechanisms, and the interventions performed in each experimental context.
Thus, \ours provides a principled Bayesian approach for propagating the epistemic uncertainty across all latent quantities
% , i.e., the causal structure, parameters, and the interventions, 
which would be infeasible when treating uncertainty quantification and intervention identification separately or sequentially.
Moreover, \ours operates in the continuous space of latent probabilistic representations of both causal Bayesian networks (CBNs) and interventions.
This formulation allows us to approximate the complex joint posterior via efficient, gradient-based particle variational inference techniques, making our approach particularly scalable to causal systems of many variables.
%, -- allowing us to reason about the {\em epistemic uncertainty} of its structural predictions. 
%
% The general formulation of \ours  makes it agnostic to the statistical structure of the local causal effects and interventions. Thus, we can instantiate our framework with common assumptions such as linear Gaussian or neural network based conditionals.

\looseness -1 In a range of experiments conducted on synthetic causal BNs and a realistic gene-expression simulator, \ours significantly outperforms related methods in recovering the underlying causal structure and intervention targets. This holds even when the underlying model is strongly misspecifed as in the case of gene-expression data.
\vspacecaption
\section{RELATED WORK}
\vspacecaptionlow
\label{sec:related}

\textbf{Continuous optimization for structure learning. }
Classical algorithms for causal structure learning typically rely on conditional independence tests \citep{spirtes2000causation, solus2017consistency} or combinatorial search \citep{chickering2002optimal, huang2018generalized}. 
Initiated by \citet{zheng2018dags}, recent works reformulate structure learning as a continuous optimization problem \citep{zheng2020learning, yu2019dag, lachapelle2019gradient, brouillard2020differentiable}, which allows using gradient-based learning algorithms for this task.
Building on these techniques, a recent line of work considers performing Bayesian inference over the causal graph instead of learning a point estimate \citep{annadani2021variational, lorch2021dibs, cundy2021bcd}. Our approach belongs to this category but is the first to consider Bayesian inference from multiple contexts with unknown interventions.

% In a recent line of work initiated by , the task of finding the optimal DAG has been reformulated as a continuous optimization problem . This allows for gradient-based learning algorithms that avoid the combinatorial space of DAGs.

\textbf{Causal discovery from multiple contexts. }
Learning a causal structure from datasets collected in different interventional contexts of the same causal system has been referred to as Joint Causal Inference
(JCI) \citep{mooij2016multiple}. Multiple methods handle such combinations of observational and interventional data \citep{magliacane2016ancestral, ijcai2017-187, hauser2012characterization, wang2017permutation, yang2018characterizing}, but assume full knowledge of the interventions. Other methods build on the notion of {\em invariance} \citep{anticausal,peters2016causal, meinshausen2016methods,ghassami2017learning,heinze2018invariant,Huangetal20b}, but either do not generalize to full graphs or make restrictive assumptions about the local causal effects.

\textbf{Causal discovery with unknown interventions. }
\citet{mooij2016multiple}, \citet{squires2020permutationbased} and \citet{Wang_Cao_Yu_Liang_2022} make the unknown intervention setting amenable to standard causal discovery tools such as conditional independence and invariance tests. However, hypothesis tests typically require large datasets, making these methods brittle for realistic datasets of small size. \citet{jaber2020unknown_targets} develop theoretical equivalence classes for a mixture of distributions with unknown interventions, focusing on the infinite sample setting.
% . However, their work focuses on soft interventions and distributions as inputs, not the finite sample setting.
%

\looseness-1
Other recent works formulate continuous relaxations of the joint inference problem under unknown interventions
and use gradient-based optimization to find the graph and targets \citep{ke2019dsdi, brouillard2020differentiable, faria2022differentiable}. However, these methods only infer point estimates. We adopt a fully Bayesian approach that enables principled uncertainty quantification. \citet{eaton2007exact} is the only work doing the same, though only for discrete variables and using a dynamic programming approach that does not scale to large graphs.
%\citet{jaber2020causal} investigate theoretical properties of learning from unknown interventions but only focus on soft interventions, similar to \citet{rothenhausler2015backshift}. 

% More recent works rely on differentiable optimization. \citet{ke2019dsdi} learn a DAG from a black-box model that gives unknown interventions, but their work is limited to point predictions. In comparison, the DCDI model \citep{brouillard2020differentiable} solves an augmented Lagrangian objective that can handle unknown interventions. However, they consider large dataset sizes and cannot provide uncertainty estimates.

% Moreover, many methods such as ICP \citep{} that rely on invariance of causal effects generalize to unknown interventions, however, they lack the power to induce full causal graphs.
\vspacecaption
\section{BACKGROUND: CAUSAL DISCOVERY}
\vspacecaptionlow
\label{sec:background}

\textbf{Causal Bayesian Networks. }
A Bayesian network (BN) $(\G, \param)$ uses a directed acyclic graph (DAG) to model the joint density $p(\rvx)$ of $d$ variables $\rvx=\evx_{1:d}$ via conditional probabilities. The joint distribution $p$ factorizes into a product of local conditional distributions $p_i(\evx_i|\evx_{\pa{i}}, \param)$ for each variable $\evx_i$,
% $
%     p(\evx_1,\dots,\evx_d|\param, \G) = \prod_{i=1}^d p_i(\evx_i|\evx_{\pa{i}}, \param)
% $
where $\pa{i}$ is the set of parents of node $i$ in $\G$ and the parameters $\param$ describe the exact local conditional distributions. In a \emph{causal} BN (CBN), the edges describe direct causal relations.
%between nodes not only describe conditional probabilities, but direct causal relations.
For causal structure learning, we assume that there are no unmeasured confounding variables  \citep[causal sufficiency, cf.][]{pearl_2009, spirtes2000causation, Peters2017}.
% For the purpose of causal inference, we assume that there are no hidden confounding variables (causal sufficiency) \citep{pearl_2009, spirtes2000causation, Peters2017}.

\textbf{Interventions. }
An intervention on the variable $\evx_i$ corresponds to replacing the conditional distribution $p_i$ by a new distribution $p_i^I$. An intervention is typically considered \emph{imperfect} (soft) if the dependence on the causal parents remains and \emph{perfect} (hard, structural) if all dependencies to the causal parents are removed, resulting in the mutilated graph $\G^{I_k}$ \citep[e.g.][]{pearl_2009, Peters2017}.
% Graphically, the latter corresponds to cutting incoming edges of each targeted node in $I_k$ which results in a \emph{mutilated graph} $\G^{I_k}$ 

\looseness -1 In this work, we assume the setting of a collection of $M$ interventions $\I:=(I_1, \dots I_M)$, where each intervention $I_k := (I^{\text{tar}}_k, \param_{I_k})$ acts on a set of targets $\Itar_k\subseteq\{1,\dots,d\}$. We use $\param_{I_k}$ to denote parameters that describe the individual conditional distributions $p_i^{I_k}(\evx_i|\evx_{\pa{i}}, \param_{I_k})$ induced by the intervention $I_k$ on the target variables $\{x_i | ~ i \in \Itar_k\}$. To simplify our exposition, we assume perfect interventions, 
%that make the target variables $x_i$ statistically independent of the other variables, 
i.e., $p_i^{I_k}(\evx_i|\evx_{\pa{i}}, \param_{I_k}) =  p_i^{I_k}(\evx_i| \param_{I_k})$. However, all the arguments made in this paper also hold for soft interventions. 
%
%Its statistical effect on the target variables $\{x_i | ~ i \in \Itar_k\}$ can be quantified by conditional (interventional) distributions $p_i^{I_k}(\evx_i|\evx_{\pa{i}}, \param_{I_k})$. By $\param_{I_k}$ we denote the parameters that describe these conditional distributions for intervention $I_k$.
The full data distribution under intervention $I_k$ factorizes into the observational and interventional conditionals:
\begin{align}
\begin{split}
    p(\rvx|\param,\G,I_k) = \prod_{i\notin I_k} p_i(\evx_i|\evx_{\pa{i}}, \param) \cdot \prod_{i\in I_k} p_i^{I_k}(\evx_i| \param_{I_k}) \nonumber
\end{split}
\end{align}
\looseness-1
The local conditional distributions of the variables that are not intervened upon do not change with respect to the observational distribution, which is often referred to as invariance \citep{peters2016causal} or modularity \citep{Peters2017}.

\textbf{Bayesian Inference of BNs. }
Given passively collected i.i.d.\ observations $\mathcal{D} = \{\rvx^{(1)}, \dots, \rvx^{(N)}\}$, Bayesian inference over BNs aims to estimate the \emph{full posterior} probability density over BNs that model the observations. Following \citet{friedman2003being}, given a prior distribution over DAGs $p(\G)$ and a prior over BN parameters $p(\param | \G)$, Bayes' Theorem yields the posterior distribution
\begin{align}
	p(\G, \param | \train) 
		&\propto p(\G) p(\param | \G) p(\train | \G, \param) \label{eq:bayesian-structure-learning-joint-posterior} % \quad \longrightarrow \quad \E_{p(\G, \param | \train)} \Big [	f(\G, \param)	\Big ]
% 		\\
% 	p(\G | \train) 
% 		&\propto p(\G) p(\train | \G) \label{eq:bayesian-structure-learning-marg-posterior} %\;,\\
\end{align}
where 
$p(\train | \G, \param) = \prod_{i=1}^n p(\rvx^{(i)} | \G, \param)$ is the likelihood of the independent observations in $\mathcal{D}$, here without interventions on the system.
% $p(\train | \G)$~$=$~$\int p(\param | \G) p(\train | \G, \param) d\param$ is the marginal likelihood. Thus, $p(\train | \G)$ in Eq. \ref{eq:bayesian-structure-learning-marg-posterior} is only available in closed form for special conjugate cases.
Given the posterior, the Bayesian formalism allows us to compute expectations of the form
\begin{align} \label{eq:bayesian-structure-learning-goal}
	\E_{p(\G, \param | \train)} \Big [	f(\G, \param)	\Big ] 
	%\quad \quad \text{or} \quad \quad
	%\E_{p(\G | \train)} \Big [f(\G)\Big ] 
\end{align}

for any function $f$ of interest. For instance, to obtain the posterior predictive, we would use $f(\G, \param) = p(\rvx | \G, \param)$ 
% or $f(\G)$~$=$~$p(\rvx | \G)$, respectively
~\citep{madigan1994model,madigan1995eliciting}. 
In active learning of CBNs, a commonly used $f$
% \vspace*{-3pt} % because of in-line math
is the expected information gain about $\G$ after certain interventions
\citep{tong2001active,murphy2001active,cho2016reconstructing,agrawal2019abcdstrategy}. 
%
% Contrary to point estimates, the Bayesian formulation allows sampling alternative graphs and parameters from the posterior that could explain the data. 
The posterior $p(\G, \param | \train)$ captures the epistemic uncertainty in the unknown graph and parameters. 
In the large sample limit, the weight of the prior vanishes and the posterior converges to the graphs and parameters that give the highest likelihood.
Inferring the posterior is computationally challenging
since there are super-exponentionally many, i.e.,  $\smash{\mathcal{O}(d! 2^{\binom{d}{2}})}$, possible DAGs with $d$ nodes \citep{robinson1973counting}.
Hence, computing the normalization constant $p(\train)$ is generally intractable.
\vspacecaption
\section{BAYESIAN CAUSAL DISCOVERY WITH MULTI-CONTEXT DATA}
\vspacecaptionlow
\label{sec:method}

\textbf{Problem Statement. }
\looseness - 1 In this section, we develop a general method for Bayesian inference of the CBN given multiple interventional datasets with unknown intervention targets and effects. 
% This setting is also known as learning from multiple contexts \cite{mooij2016multiple}.
%
Formally, we are given $M$ datasets $\mathbf{D} = \{\train_1, ..., \train_M\}$ that were generated under (unknown) interventions $\I=\{I_1,\dots,I_M\}$. Each $\train_k$ is a set of i.i.d.\ samples $\train_k=\{\rvx^{(k,1)},\dots,\rvx^{(k,n_k)}\}$ obtained from the interventional data distribution $p(\rvx|\param,\G,I_k)$. Observational data, if available, can be added to $\mathbf{D}$ as $\train_0 := \train$ with intervention targets $I_0 = \emptyset$.
%When we also have observation data $\train$, we can add it to $\mathbf{D}$ as $\train_0 := \train$ and set the corresponding intervention targets to the empty set $I_0 = \emptyset$.

\looseness-1
Our goal is to infer the full, unknown CBN $(\G, \param)$ and the unknown interventions $\mathcal{I}$ given the datasets $\mathbf{D}$. The key difficulty compared to standard causal inference is that, in addition to the ground truth CBN, the intervention targets $I^{\text{tar}}_k$ and their data-generating parameters $\param_{I_k}$ are unknown. Therefore, we effectively perform joint inference over $M$ mutilated graphs that are all connected through the unobserved graph $\G$.

\looseness-1
Such inference is well-posed only if the structural changes implied by each intervention $I_k$ are sparse relative to the overall size of $\G$, i.e. $|I_k| \ll d$. If large parts of the causal model were intervened upon differently across contexts, there may be little or no overlap in the mutilated causal DAGs and thus few reasons for multitasking. This assumption has also been used in prior works \citep{brouillard2020differentiable} and is commonly referred to as the Sparse Mechanism Shift hypothesis \citep{scholkopf2021toward}, which posits that distribution changes tend to manifest themselves in a sparse way.

% Our goal is to infer the ground truth CBN $(\G_{\text{gt}}, \param_{\text{gt}})$ as well as the interventions ${I_1, ..., I_k}$ based on the available data $\mathbf{D}$.
% %
% Compared to standard causal inference (TOOD: CITE STH!!), the key difficulty is that, in addition to the ground truth CBN, both the {\em intervention targets} (i.e. $I^{\text{tar}}_k$) and the {\em statistical effect} of each intervention (i.e. $\param_{I_k}$) are unknown. So overall, we have data structured in $M$ 'buckets' and need perform joint inference over the $M$ mutilated graphs which are all closely connected to a unknown prototype graph. Naturally, performing such joint inference is only fruitful if the structural changes implied by each intervention $I_k$ are sparse compared to the overall size of the causal graph $\G_{\text{gt}}$, i.e., $|I_k| \ll d$. 
\looseness-1
\textbf{Bayesian inference.} In many relevant application domains such as biology, the observed samples $n_k$ are noisy and small in number. Hence, it is paramount to not only to predict a single prototype CBN alongside one intervention hypothesis per dataset but also to reason about their joint  epistemic uncertainty. Such uncertainty estimates allow us to quantify the reliability of our predictions and can be used to actively design future experiments. 
%In addition, they can also be used to actively design informative intervention experiments to help reduce the epistemic uncertainty further.
%
We thus approach the problem from a Bayesian perspective. This renders the task of learning from $\mathbf{D}$ as a posterior inference problem, where the prior probability and likelihood function are chosen beforehand and serve as the antecedents of inference. We are going to gradually build up this problem in the following.

%To this end, we build upon recent advances for approximate posterior inference in Bayesian Structure Learning made by the DiBS framework \citep{lorch2021dibs}. In its current form, DiBS only infers the posterior distribution given purely observational data and thus generally cannot distinguish between Markov equivalent graphs without additional assumptions \citep{pearl_2009}.

\textbf{Known interventions.} When the intervention targets $I^{\text{tar}}_k$ and the parameters $\param_{I_k}$ of the intervention effect are known, for $k=1,\hdots,M$, the posterior over CBNs includes \begin{enumerate*}[label=(\roman*.)] \item the product of data likelihoods over all datasets in $\mathbf{D}$, and \item interventional instead of observation likelihoods for $\train_{k \geq 1}$\end{enumerate*},

\begin{align}
\begin{split}
    p(\G, \param|\mathbf{D}, \I) & \propto{} \underbrace{p(\G)p(\param|\G)}_{\text{priors}} \underbrace{p(\mathcal{D}_0|\param, \G)}_{\text{obs. likelihood}}  \\ 
    &~~~~ \cdot \prod_{k=1}^M \underbrace{p(\mathcal{D}_k|\param, \G, I_k)}_{\text{interv. likelihood}} \label{eq:posterior_known_interventions}\;,
    \vspaceequation
\end{split}
\end{align}

where $p(\mathcal{D}_k|\param, \G, I_k) = \prod_{i=1}^{n_k} p(\rvx^{(k, i)}|\param,\G,I_k)$ is the interventional likelihood for $\mathcal{D}_k$ given $I_k = (I^{\text{tar}}_k, \param_{I_k})$.

\textbf{Unknown interventions.} We include unknown interventions in our inference model by introducing additional priors $p(I^\text{tar}_k)$ and $p(\param_{I_k} | I^\text{tar}_k)$. Accordingly, the modified posterior follows as
\begin{align}
\begin{split}
    p(\G, \param, \I| & \mathbf{D}) \propto{}  \underbrace{p(\G)p(\param|\G)}_{\text{priors}} \underbrace{p(\mathcal{D}_0|\param, \G)}_{\text{obs. likelihood}} \\ 
    & \cdot \prod_{k=1}^M \underbrace{p(I^{\text{tar}}_k)p(\param_{I_k}|I^{\text{tar}}_k)}_{\text{interv. priors}} \underbrace{p(\mathcal{D}_k|\param, \G, I_k)}_{\text{interv. likelihood}} \label{eq:posterior_unknown_interventions}
    \vspaceequation
\end{split}
\end{align}
The prior $p(I^{\text{tar}}_k)$ over intervention targets models prior beliefs about the structure of interventions, e.g., that only a sparse set of variables are subject to an intervention at the same time. The parameterization of the interventional distributions $p_i^{I}(\evx_i| \param_{I})$ is informed by the application and reflects the general nature of interventions, e.g., gene knockdowns in biology.

\vspacecaption
\section{A DIFFERENTIABLE GENERATIVE MODEL OVER CBNs AND INTERVENTIONS}
\vspacecaptionlow
\label{sec:diff_model_over_cbs_interv}
\looseness-1
In the following, we represent $\G \in \{0, 1\}^{d \times d}$ as the adjacency matrix and $\Itar_k = [\Itar_{k,1}, ..., \Itar_{k,d}]^\top \in \{0, 1\}^d$ as the indicator vector where $\Itar_{k,l} = 1$ if the $l$-th variable is intervened upon and $\Itar_{k,l} = 0$ otherwise. 
Since multiple nodes may be intervened upon simultaneously,
$\Itar_{k}$ is in general not one-hot.
We write $\I^{\text{tar}}$ short for the stack $[I^{\text{tar}}_1, ...,  I^{\text{tar}}_M]$ of intervention targets and $\param_\I := [ \param_{I_1}, ..., \param_{I_M} ]$ for the intervention effect parameters.
% As discussed in Sec. \ref{sec:background}, the posterior in Eq. \ref{eq:posterior_unknown_interventions} is highly intractable as the number of possible DAGs $\G$ grows super-exponentially with the number of variables $d$. Furthermore, the number of possible intervention targets $I^{\text{tar}}_k$ %\subseteq \{1, ..., d\}$
% grows in the order of $\mathcal{O}(2^d)$. 

% To facilitate approximate inference of the posterior in Eq.~\ref{eq:posterior_unknown_interventions}, we harness recent advances in Bayesian structure learning proposed by \citet{lorch2021dibs}. % that allow for a {\em fully differentiable} posterior over graphs. 
% Specifically, by allowing the approximation of the score of the posterior, their extended generative model enables the use of efficient Bayesian inference methods such as variational inference \citep{blei2017} and Stein Variational Gradient Descent (SVGD) \citep{liu2016stein}.

\textbf{Challenges.} The Bayesian inference task is intricate when learning from multiple datasets generated under interventions with unknown targets and effects.
Learning the joint posterior %over the CBN and interventions 
in Eq.~\ref{eq:posterior_unknown_interventions} requires working with a complex distribution over discrete DAGs $\G$, continuous mechanism parameters $\param$, and interventions $\smash{\{I_k\}_{k=1}^M}$, which in turn affect the identification of the DAG $\G$ itself.
Consequently, alternating inference of $\G$ and $\I^{\text{tar}}$ using an EM-like approach would preclude propagation of epistemic uncertainty across all latent quantities and thus lead to sub-optimal results.

Moreover, it is essential to infer the parameters of the interventional likelihoods $p_i^{I_k}(\evx_i| \param_{I_k})$ in Eq.~\ref{eq:likelihood} {\em conditional upon} predicting that an intervention occurred. This is of particular importance when interventions constitute a strong shift of distributions. By naively masking the observational likelihood when a variable is believed to be targeted, we would not evaluate the likelihood of the intervention itself and effectively operate outside the Bayesian framework. This would encourage the prediction of interventions whenever our learned model is suboptimal in explaining the data.

\looseness-1
To tackle these joint inference challenges,
we utilize ideas of \citet{lorch2021dibs}, who introduce a method for efficient inference of the posterior of the CBN ($\G, \param$) given a single observational dataset $\D$ by translating the distribution into a continuous latent space.
Generalizing their approach, we transform our multi-context inference problem over the DAG $\G$ and the set of interventions targets $\{\Itar_k\}$ into one over only continuous latent variables that is {\em consistent} with the original task in Eq.~\ref{eq:posterior_unknown_interventions} and that allows for the direct estimation of the score of the {\em joint} posterior over $\G$ and $\Itar_k$ in each context $k=1, \dots, M$.
Devising such an inference scheme for the multi-context, unknown intervention setting requires careful modeling of the intervention target prior that accurately captures our assumptions of the data generating process, such as sparsity and sharpness of the interventions. At the same time, it must enable accurate and tractable inference via methods like variational inference \citep{blei2017} and perform well in practice.
% A major task for devising the inference scheme for the multi-environment, unknown intervention setting is careful modeling of the intervention target prior as well as likelihood. We require the model to be both accurate and tractable. At the same time, it needs to accurately capture our assumptions of the data generating process such as sparsity and sharpness of the interventions. 

To enable joint inference of all latent quantities, we introduce continuous latent variables $\Z$ and $\intvGam_k$ and their corresponding priors, which model the generative processes of $\G$ and $\Itar_k$ through $p(\G|\Z)$ and $p(\Itar_k | \intvGam_k)$. 
This implies the following extended factorization of the generative model, which is also given in Figure~\ref{fig:generative_model}:
\begin{figure}
    \centering
    % \begin{tikzpicture}[x=1.0cm,y=0.5cm]
%   % Define nodes
%   \node[latent] (gamma) {$\intvGam_{k}$};
%   \node[latent, right=of gamma, xshift=-0.35cm] (target) {$\I_{k}^{\text{tar}}$};
%   \node[latent, right=of target]  (thetatarget) {$\param_{I_k}$};
%   \node[obs, below=of target, xshift=0.9cm, yshift=0.1cm] (x)
%   {$\rvx$};
%   \node[const, above=of x, yshift=-0.4cm]  (dummy) {$~$}; % dummy for controlling the height of inner box

%   \node[latent, below=of x, xshift=-0.9cm, yshift=-0.35cm] (g) {$\G$};
%   \node[latent, left=of g, xshift=0.35cm] (z) {$\Z$};
%   \node[latent, right=of g]  (theta) {$\param$};

% %  \node[const, left=of z] (beta) {$\beta$};
% %  \node[const, above=of beta, yshift=0.3cm] (alpha) {$\alpha$};

%   % Connect the nodes
%   \edge {gamma} {target} ; %
%   \edge {target} {thetatarget} ; %
%   \edge {target, thetatarget} {x} ; %

%   \edge {z} {g} ; %
%   \edge {g} {theta} ; %
%   \edge {g, theta} {x} ; %

% %   \edge {beta} {z} ; %
% %   \edge {alpha} {g} ; %
% %   \edge {alpha} {target} ; %

%   % Plates
%   {
% %   \tikzset{plate caption/.append style={below right=9pt and 0pt of #1.south west}} % moves location of plate label
%   \tikzset{plate caption/.append style={below left=9pt and 0pt of #1.south east}} % moves location of plate label
%   \plate {} {(gamma) (target) (thetatarget) (x)} {$M$} ;
%   }
%   {
%   \tikzset{plate caption/.append style={below right=5pt and 6pt of #1.south west}} % moves location of plate label
%   \plate {} {(dummy) (x)} {$n_k$} ;
%   }
% \end{tikzpicture}

%% version 2

\begin{tikzpicture}[x=1.0cm,y=0.5cm]
  % Define nodes
  \node[latent] (gamma) {$\intvGam_{k}$};
  \node[latent, right=of gamma, xshift=-0.3cm] (target) {$I_{k}^{\text{tar}}$};
  \node[latent, below=of target]  (thetatarget) {$\param_{I_k}$};
  \node[obs, right=of target, xshift=0.0cm, yshift=-0.6cm] (x)
  {$\rvx$};
  \node[const, right=of x, xshift=-0.8cm, yshift=-0.0cm]  (dummy) {$~$}; % dummy for controlling the size of bigger box

  \node[latent, right=of x, xshift=-0.0cm, yshift=0.6cm] (g) {$\G$};
  \node[latent, right=of g, xshift=-0.3cm] (z) {$\Z$};
  \node[latent, below=of g]  (theta) {$\param$};

%  \node[const, left=of z] (beta) {$\beta$};
%  \node[const, above=of beta, yshift=0.3cm] (alpha) {$\alpha$};

  % Connect the nodes
  \edge {gamma} {target} ; %
  \edge {target} {thetatarget} ; %
  \edge {target, thetatarget} {x} ; %

  \edge {z} {g} ; %
  \edge {g} {theta} ; %
  \edge {g, theta} {x} ; %

%   \edge {beta} {z} ; %
%   \edge {alpha} {g} ; %
%   \edge {alpha} {target} ; %

  % Plates
  {
%   \tikzset{plate caption/.append style={below right=9pt and 0pt of #1.south west}} % moves location of plate label
  \tikzset{plate caption/.append style={below left=0pt and 0pt of #1.south east}} % moves location of plate label
  \plate {} {(gamma) (target) (thetatarget) (x) (dummy)} {$M$} ;
  }
  {
%   \tikzset{plate caption/.append style={below right=5pt and 6pt of #1.south west}} % moves location of plate label
%   \plate {} {(dummy) (x)} {$n_k$} ;
  \plate {} {(x)} {$n_k$} ;
  }

\end{tikzpicture}
    \caption{
    \looseness-1
    Generative model of causal Bayesian networks with observations sampled in $M$ intervention contexts.
    The continuous variables $\{\intvGam_k\}$ and $\Z$ extend the default data-generating process and allow reformulating the Bayesian inference task for gradient-based inference techniques.
    }
    \vspacecaption
    \label{fig:generative_model}
\end{figure}
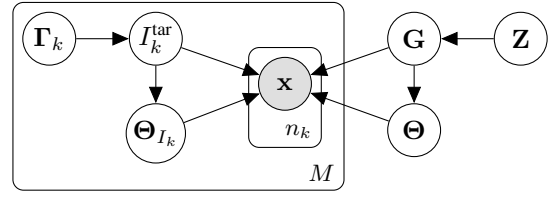
%
% The key challenge is that differentiating w.r.t. discrete structures $\G$ and $\Itar_k$ is not possible. For this reason, we introduce continuous latent variables $\Z$ and $\intvGam_k$ for $k=1, ..., M$ that model the generative processes of $\G$ and $\Itar_k$ through $p(\G|\Z)$ and $p(\Itar_k|\intvGam_k)$. This allows us to transform the inference problem over discrete structures into one over continuous parameters which we can differentiate w.r.t.. This implies the following factorization of the joint probability of the generative model in Figure~\ref{fig:generative_model}:
\vspaceequation
\begin{align} \label{eq:data_gen_process}
    & p(\Z, \G, \param, \intvGam, \I, \mathbf{D}) = \underbrace{p(\Z)p(\G|\Z) p(\param|\G)}_{\text{generative process CBN}} \\
    & \cdot \prod_{k=1}^{M} \underbrace{p(\intvGam_k)p(\Itar_k|\intvGam_k) p(\param_{I_k} | \Itar_k)}_{\text{generative process intervention}} \underbrace{p(\mathcal{D}_k|\G, \param, \Itar_k, \param_{I_k})}_{\text{interventional likelihood}} \notag \vspace{-4pt}
\end{align}
where $\intvGam := [\intvGam_1,...,\intvGam_M]$ for brevity. 
As shown in the following, the extended generative model we introduce allows expressing the posterior in Eq.~\ref{eq:posterior_unknown_interventions} in terms of the posterior over the continuous latent variables  $\Z$, $\param$, $\intvGam$, and $\param_\I$:
\begin{proposition}
\label{prop:latent_post_expectation}
    Under the extended generative model in Eq.~\ref{eq:data_gen_process} and Figure~\ref{fig:generative_model}, it holds that
\begin{alignat}{2}
 \label{eq:latent_post_expectation}
     & \E_{p(\G, \param, \I | \mathbf{D})} [f(\G, \param, \I)] =  \\
     & \E_{p(\Z, \param, \intvGam, \param_\I |  \mathbf{D})} \left[\frac{\E_{p(\G|\Z)}\E_{p(\I^{\text{\normalfont tar}}|\intvGam)}[f(\G, \param, \I)\cdot \mathbf{\Psi}]}{\E_{p(\G|\Z)}\E_{p(\I^{\text{\normalfont tar}}|\intvGam)}[\mathbf{\Psi}]} \right] \notag 
\end{alignat}
with weighting $\mathbf{\Psi}=p(\param|\G)p(\param_\I|\I^{\text{\normalfont tar}})p(\mathbf{D}|\G,\I,\param)$ and $p(\mathbf{D}|\G,\I,\param)= \prod_{k=1}^{M} p(\mathcal{D}_k|\G, \param, I^{\text{\normalfont tar}}_k,\param_{I_k})$.
% \begin{alignat}{2}
%      &p(\G, \param, \I | \mathbf{D}) =  \iint p(\Z, \param, \intvGam, \param_\I |  \mathbf{D}) \, \Psi\, d\Z d\intvGam \notag \\
%      &\text{where }\Psi= 
%      \frac
%         {p(\G|\Z)p(\I^{\text{\normalfont tar}}|\intvGam) p(\param_\I|\I^{\text{\normalfont tar}})p(\param,\mathbf{D}|\G,\I)}
%         {\E_{p(\G^\prime|\Z)}\E_{p(\I^{\text{\normalfont tar$\prime$}}|\intvGam)}\big[p(\param_\I|\I^{\text{\normalfont tar$\prime$}})p(\param,\mathbf{D}|\G^\prime,\I^\prime)\big]}\hspace*{-1pt}
%      \hspace*{-2pt}\notag 
% \end{alignat}
% and $p(\param,\mathbf{D}|\G,\I) = p(\param|\G)p(\mathbf{D}|\G,\I,\param)$.
% \begin{alignat}{2}
% %  \label{eq:latent_post_expectation}
%      & \E_{p(\G, \param, \I | \mathbf{D})} [f(\G, \param, \I)] =  \\
%      & \E_{p(\Z, \param, \intvGam, \param_\I |  \mathbf{D})} \left[\frac{\E_{p(\G|\Z),p(\I^{\text{\normalfont tar}}|\intvGam)}[f(\G, \param, \I)\cdot \mathbf{H}]}{\E_{p(\G|\Z),p(\I^{\text{\normalfont tar}}|\intvGam)}[\mathbf{H}]} \right] \notag 
% \end{alignat}
% %with $p(\mathbf{D}|\G,\I,\param) = \prod_{k=1}^{M} p(\mathcal{D}_k|\G, \param, \I^{\text{\normalfont tar}}_k,\param_{I_k})$.
\end{proposition}
% and $p(\I|\intvGam) = \prod_{k=1}^{M} p(\Itar_k|\intvGam_k) p(\param_{I_k}| \Itar_k)$.
% This theorem states that the Bayesian expectation over discrete structures can be expressed with the continuous parameters $\Z, \param, \intvGam, \param_\I$. 
This insight shows that the posterior expectation over graphs and interventions can be computed via an expectation over the latent posterior $p(\Z, \param, \intvGam, \param_\I |  \mathbf{D})$. 
We provide a proof in Appx. \ref{appx:proof_latent_posterior_expectation}.
The inner term is akin to a likelihood ratio of the considered $(\G, \param, \mathcal{I})$ under the posterior expectation over $\Z$ and $\intvGam$. 
All factors besides the latent posterior are tractable to compute or approximate.
Before we discuss how to perform approximate inference of $p(\Z, \param, \intvGam, \param_\I |  \mathbf{D})$, we further specify the conditional probabilities of our generative model and how to make them differentiable.

\textbf{Generative model of DAGs $\G$.} Following \citet{lorch2021dibs}, we define the latent variable $\Z$ as the stack of embedding matrices $\mathbf{U},\mathbf{V} \in \R^{d \times d}$ and the generative model for the adjacency matrix $\G$ by using the inner product:
\vspaceequation
\begin{align}
\begin{split}
    p_\alpha(\G| \Z) = & \prod_{i=1}^d \prod_{j \neq i}^d p_\alpha(g_{ij} | \mathbf{u}_i, \mathbf{v}_j) \\
	~~~
	\text{with}
	~~~
	& g_{ij} |  \mathbf{u}_i, \mathbf{v}_j \sim \text{Bern}( \sigma_\alpha(\mathbf{u}_i^\top \mathbf{v}_j))
	\vspaceequation
\end{split}
\end{align}
where $\sigma_\alpha(x) = 1/(1 + \exp(-\alpha x))$ is the sigmoid function with inverse temperature $\alpha$ and $\mathbf{u}_i$, $\mathbf{v}_j$ the $i$-th and $j$-th column vectors of $\mathbf{U}$ and $\mathbf{V}$, respectively. The authors show that this formulation outperforms a scalar parametrization based on a $d\times d$ matrix. We denote the matrix of edge probabilities in $\G$ given $\Z$ by $\G_\alpha(\Z) \in [0, 1]^{d \times d}$ with $\G_\alpha(\Z)_{ij}:= \sigma_\alpha(\mathbf{u}_i^\top \mathbf{v}_j)$. 
We model the prior over $\Z$ as \begin{enumerate*}[label=(\roman*.)] \item i.i.d.~Gaussian with a variance of $\eta_{Z}^2 = 1/d$ to ensure well-behaved gradients and \item an acyclicity term that penalizes the expected cyclicity of $\G$ given $\Z$ \end{enumerate*}:
\vspaceequation
\begin{align} \label{eq:dibs-dag-model}
\begin{split}
    p_{\beta}(\mathbf{Z}) = p(\mathbf{U}, \mathbf{V}) & \propto \underbrace{\exp \left( - \beta \E_{p(\G|\Z)} [ h(\G)] \right) }_{\text{acyclicity prior}} \\
    & \cdot \prod_{i=1}^d \underbrace{\mathcal{N}(\mathbf{u}_i| \boldsymbol{0}, \eta_{Z}^2 \mathbf{I}) \mathcal{N}(\mathbf{v}_i| \boldsymbol{0}, \eta_{Z}^2 \mathbf{I})}_{\text{inference stability}} \;
    \vspaceequation
\end{split}
\end{align}
Here, $\beta$ is an inverse temperature parameter controlling how strongly the acyclicity is enforced, and $h(\G) = \text{tr} \left[ (I + \frac{1}{d} \G)^d \right] - d \geq 0$. By Theorem 1 of \citet{yu2019dag}, $\G$ is acyclic iff $h(G) = 0$. As $\beta \rightarrow \infty$, the support of $p(\mathbf{Z})$ reduces to all $\Z$ that model DAGs.

\textbf{Generative model of intervention targets $\I^{\text{tar}}$.} 
% Similar to the generative model for $\G$, 
To model the intervention targets in continuous space,
we introduce the latent variable $\intvGam \in \R^{M\times d}$.
Each $\gamma_{k,i}$ is the logit of an independent Bernoulli distribution and models one entry $(k,i)$ of the intervention target mask $\I^\text{tar} = [\Itar_1, ..., \Itar_M] \in \{0, 1\}^{M \times d}$:
\vspaceequation
\begin{align}\label{eq:intervention-model}
\begin{split}
    p(\I^\text{tar} | \intvGam) =& \prod_{k=1}^M \prod_{i=1}^d p_\alpha(\Itar_{k,i} | \gamma_{k, i}) \\
	~~~
	\text{with}
	~~~
	& \Itar_{k,i} |  \gamma_{k, i} \sim \text{Bern}(\sigma_\alpha(\gamma_{k, i})) 
	\vspaceequation
\end{split}
\end{align}
\looseness -1 We denote the matrix of intervention target probabilities as $\I^\text{tar}_\alpha(\intvGam) \in [0, 1]^{M \times d}$ with $\I^\text{tar}_\alpha(\intvGam)_{k,i} = \sigma_\alpha(\gamma_{k, i})$. 
% Similar to $\Z$,
The prior over $\intvGam$ has three components: \begin{enumerate*}[label=(\roman*.)] \item A Gaussian term for {\em inference stability} , \item a Beta-distribution {\em sharpness} prior
% $\text{Beta}(\sigma_{\alpha}(\gamma_{k,i}); \zeta_1, \zeta_2)$
that encourages $\sigma_{\alpha}(\gamma_{k,i})$ to be close to 0 or 1%(whether variable $i$ in context $k$ is intervened on)
, and \item a {\em sparsity} prior
% $p_\text{sparsity}(\gamma_{k}) = \exp \left( -\lambda \lVert \sigma_{\alpha}(\gamma_{k}) \rVert_{1} \right)$
with the $l_1$-norm of $\sigma_{\alpha}(\intvGam_{k})$ and the inverse temperature parameter $\lambda$. \end{enumerate*}
\vspaceequation
\begin{align}
\begin{split}
     p(\intvGam) \propto & \prod_{k=1}^M \underbrace{\exp \left( -\lambda \lVert \sigma_{\alpha}(\intvGam_{k})  \rVert_{1} \right)}_{\text{sparse masks}}  \\
     & \cdot \prod_{i=1}^d \underbrace{\text{Beta}(\sigma_{\alpha}(\gamma_{k,i}); \zeta_1, \zeta_2)}_{\text{sharp masks}} \underbrace{\mathcal{N} (\gamma_k | \boldsymbol{0}, \eta_{\gamma}^2 \mathbf{I})}_{\text{inference stability}} \; \label{eq:prior_interv_gamma}
     \vspaceequation
\end{split}
\end{align}
\looseness -1 We assume that interventions occur infrequently, i.e., in expectation only on one variable. Hence, we choose $\zeta_1 = 1/ d$ and $\zeta_2 = (d-1) / d$. The sparsity prior implies that, given an active intervention target $i$, it is a-priori less likely that a variable $j \neq i$ is intervened upon in the same context $k$. The sparsity of interventions can additionally be regulated with the parameter $\lambda$.

\looseness -1 \textbf{Interventional likelihood.} 
Combining the generative DAG and intervention models in Eqs. \ref{eq:dibs-dag-model} and \ref{eq:intervention-model},
we obtain a differentiable interventional likelihood by sampling the graph $\G \sim \text{Bern}\left(\smash{\sigma_\alpha(\mathbf{U}\mathbf{V}^\top})\right)$ and masks $\I^{\text{tar}} \sim \text{Bern}\left(\sigma_\alpha(\intvGam)\right)$ with the Gumbel-Softmax trick \citep{jang2016categorical, maddison2017concrete} and using them to select between the observational and interventional likelihoods for each variable:
\vspaceequation
\begin{align}
  \label{eq:likelihood}
  & p(\train_k| \G, \param, \Itar_k, \param_{I_k}) = \\
  &\prod_{j=1}^{n_k} \prod_{i=1}^d \left( p(x^{(k,j)}_i | \evx_{\pa{i}}, \param)^{(1-\Itar_{k,i})} \cdot p(x^{(k,j)}_i | \param_{I_k})^{\Itar_{k,i} } \right) \notag
  \vspaceequation \vspace{-6pt}
\end{align}

Importantly, this formulation does not assume a specific likelihood or intervention model. This means that any restricted model (e.g. linear mechanisms, non-Gaussian noise) can be used to describe the causal relations between variables. This also extends to the interventions, where both hard and soft interventions can be plugged into the interventional likelihood in Eq. \ref{eq:likelihood} to model local changes in the distribution. For both parts, the only required property is differentiability with respect to the latent parameters. The likelihood model should be informed by the application, experts, or the type of data that is assumed; additionally, restricted models are necessary in order to guarantee identifiability of the ground-truth structure. We will come back to the question of identifiability at the end of Sec. \ref{sec:svgd_instantiation}.
 
% The Gumbel-softmax trick allows us to estimate gradients of the posterior with respect to all continuous latent variables.
%
% In a similar fashion, we use the $\I^{\text{tar}}$ to select between observational and interventional likelihoods per variable:
% \begin{equation*}
%   \log p(\train_k| \G, \param, \Itar_k, \param_{I_k}) \hspace{-2pt} = \hspace{-2pt} \sum_{j=1}^{n_k} \sum_{i=1}^d \left(\hspace{-2pt} (1-\Itar_{k,i}) \cdot \log p(x^{(k,j)}_i | \evx_{\pa{i}}, \param)  + \Itar_{k,i} \cdot \log p(x^{(k,j)}_i | \param_{I_k}) \hspace{-2pt}\right)
% \end{equation*}
%
%%%%%%%%%%%%%%%%%%%%%%%%%%%%%%%%%%%%%%%%%%%%%%%%%%%%%%%%%%%%%%%%%%%%%%%
\begin{figure*}[ht!]
    \centering
    \includegraphics[width=.95\linewidth]{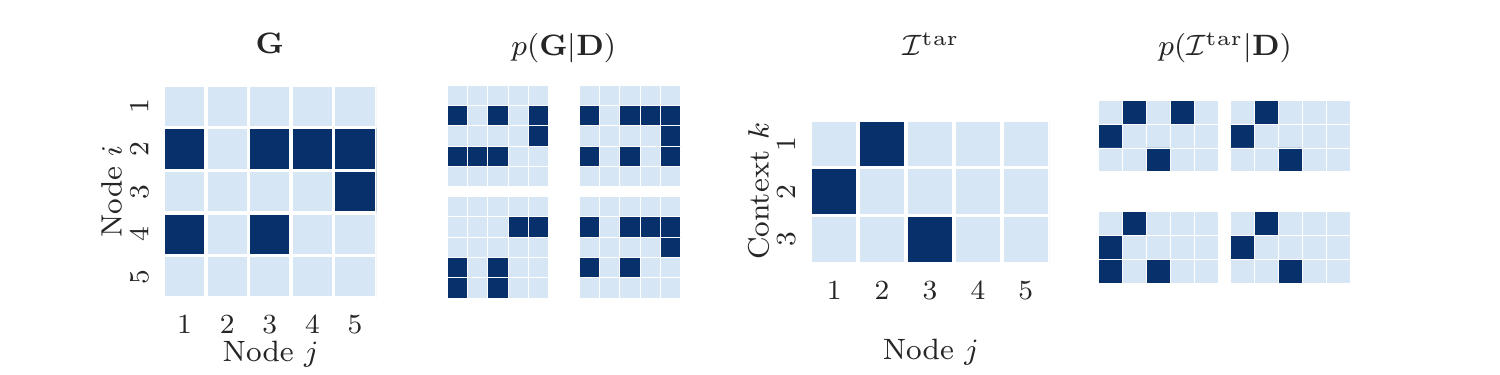}
    \vspacecaption
    \caption{\looseness -1 Instead of just one point estimate, \ours  yields particle approximations of the posteriors $p(\G | \mathbf{D})$ and $p(\I^{\text{tar}} | \mathbf{D})$. We visually compare the posterior particles with the ground truth $\G$ and $\I^{\text{tar}}$ for a SF-2 graph with $d=5$ nodes and $M=3$ contexts. Blue colors represent edges or targets.
    \vspacefigcaptionmid
    }
    %Visual comparison of samples from \ours's inferred posteriors $p(\G | \mathbf{D})$ and $p(\I^{\text{tar}} | \mathbf{D})$ with corresponding ground truths $\G_{\text{gt}}$ and $\I^{\text{tar}}_{\text{gt}}$. Blue colors represent presence of edges or targets. Here, we use a SF-2 graph with $d=5$ nodes and $M=3$ contexts.}
    \label{fig:toy_example}
\end{figure*}
\section{VARIATIONAL INFERENCE OVER CBNs AND INTERVENTIONS}
\vspacecaptionlow
\label{sec:svgd_instantiation}

In Section \ref{sec:diff_model_over_cbs_interv}, we have presented an extended graphical model of the multi-context, unknown-intervention causal discovery problem.
Our extended model introduces additional continuous variables $\Z$ and $\intvGam$ that characterize the generative process of $\G$ and the intervention targets $\I^{\text{tar}}$.
Proposition \ref{prop:latent_post_expectation} shows that we can perform Bayesian inference of the graph $\G$ and intervention targets $\I^{\text{tar}}$ by inferring the posterior over the continuous latent variables $\Z$ and $\intvGam$. 

In this section, we discuss how to approach the final challenge of inferring this equally intractable, yet continuous posterior and finally convert the result back into an approximation of the joint posterior $p(\G, \param, \I^{\text{tar}}, \param_{\I} | \mathbf{D})$ over discrete structures $\G, \I^{\text{tar}}$ and continuous parameters $\param,  \param_{\I}$.

\looseness -1 \textbf{Posterior scores.}\quad
While Proposition \ref{prop:latent_post_expectation}
links the semi-continuous posterior $p(\G, \param, \I^{\text{tar}}, \param_{\I} | \mathbf{D})$ to
its continuous counterpart, the inference problem appears just to shift.
However, the transformation we introduce enables using inference methods based on continuous optimization, and more importantly, allows estimating the gradients of $\log p(\Z, \param, \intvGam, \param_\I |  \mathbf{D})$, which many approximate inference techniques rely on \citep[e.g.][]{welling2011bayesian, chen2014stochastic,liu2016stein}.
The gradient with respect to $\intvGam$ is given by
\begin{align}
\begin{split}
    &\nabla_{\intvGam} \log p(\Z, \param, \intvGam, \param_{\I} | \mathbf{D})  \\
    &= \nabla_{\intvGam} \log p(\intvGam) +\frac{\nabla_{\intvGam} \E_{p(\G | \Z)} \E_{p(\I^{\text{tar}} | \intvGam)}\left[   \mathbf{\Psi}\right]}{\E_{p(\G | \Z)} \E_{p(\I^{\text{tar}} | \intvGam)}\left[   \mathbf{\Psi}\right]}
\end{split}
\end{align}
where $\mathbf{\Psi}$ the same as in Proposition~\ref{prop:latent_post_expectation} and contains the priors and likelihood in Eq.~\ref{eq:likelihood}. 
By sampling $\G$ and $\I^\text{tar}$ using the Gumbel-softmax trick, %(cf.\ Section \ref{sec:diff_model_over_cbs_interv}), 
we can pull the gradient $\nabla_\intvGam$ inside the expectations and tractably approximate the score using Monte Carlo sampling.
The gradients for $\Z$, $\param$, and $ \param_{\I}$ are analogous. The derivations are given in Appendix \ref{appx:scores}.
% Using the Gumbel-Softmax trick \citep{jang2016categorical, maddison2017concrete}, we can backpropagate through the expectations over $p(\G | \Z)$ and $p(\I^{\text{tar}} | \intvGam)$.

\looseness -1 \textbf{Consistent particle variational inference.}\quad
To infer the latent posterior $p(\Z, \param, \intvGam, \param_{\I} | \mathbf{D})$, we employ the particle variational inference approach Stein Variational Gradient Descent (SVGD) \citep{liu2016stein}.
SVGD minimizes the KL divergence to the intractable distribution of interest using a finite, optimized set of particles. 
Specifically, the algorithm uses the score of the density to transport particles towards high-probability regions and a kernel $k(\cdot, \cdot)$ to introduce a repulsive force between the particles. 
Since BaCaDI allows estimating the score, we can apply SVGD off-the-shelf.
We give an overview of SVGD in Appx.~\ref{appx:svgd}.

% \looseness-1 We employ a sum of RBF kernels as the kernel function over $\Psib:=(\Z, \param, \intvGam, \param_{\I})$, which performed better than a product kernel composition (see Appx.~\ref{appx:svgd_kernel}).
Given $L$ initial particles $\{ ( \Z^{(l)}_0, \intvGam^{(l)}_0, \param^{(l)}_0, \param^{(l)}_{\I, 0} )\}_{l=1}^L$, we perform $T$ iterations of particle SVGD updates. We use annealing schedules $\alpha_t \rightarrow \infty$ and $\beta_t \rightarrow \infty$ for our continuous relaxations $\G_\alpha(\Z)$, $\I_\alpha^\text{tar}(\intvGam)$ and the graph prior $p_\beta(\Z)$. 
By annealing the temperature parameters, our posterior over the continuous latent variables $\Z$ and $\intvGam$ asymptotically converges into a probability distribution over discrete DAGs and interventions (proof in Appendix \ref{appx:annealing}).:

%to converge to DAGs and discrete sets of intervention targets (Appx~\ref{appx:annealing}). 
%We return $\{ (\G_\infty (\Z^{(l)}_T), \param^{(l)}_T, \I^{\text{tar}}_\infty (\intvGam^{(l)}_{T}),  \param^{(l)}_{\I, T})  \}_{l=1}^L$ as the particle approximation of the posterior
%$p(\G, \param, \I_{\text{tar}}, \param_\I| \mathbf{D})$ 
%with discrete DAGs and interventions targets.

\begin{proposition}
\label{prop:annealed_posterior}
As $\alpha \rightarrow \infty$ and $\beta \rightarrow \infty$
the posterior expectation in Prop. \ref{prop:latent_post_expectation} converges to the simpler expression
\begin{align} \label{eq:asymptotic_expectation}
\begin{split}
    \E_{p(\G, \param, \I^{\text{\normalfont  tar}}, \param_\I | \mathbf{D})} & [f(\G, \param, \I^{\text{\normalfont tar}}, \param_\I )] \\  
     \rightarrow
 \E_{p(\Z, \param, \intvGam, \param_\I |  \mathbf{D})} & [f(\G_\infty(\Z), \param, \I_\infty^\text{\normalfont  tar}(\intvGam), \param_\I )]\
\end{split}
\end{align}
with $\G_\infty(\Z)_{i,j} = \mathbf{1}_{\mathbf{u}_i^\top \mathbf{v}_j > 0}$ and $\I_\infty^\text{\normalfont  tar}(\intvGam)_{k,i} = \mathbf{1}_{\gamma_{i, k} > 0}$. 
In the limit, the marginal posterior over discrete structures $p(\G, \I^\text{\normalfont  tar} | \mathbf{D})$ is supported on $\{\mathbf{G} | \mathbf{G} \in \{0,1\}^{d \times d} \wedge \mathbf{G} \text{ is acyclic} \} \times \{0,1\}^{M \times d}$ and, thus, a valid probability mass function over DAGs and intervention targets.
\end{proposition}

Proposition \ref{prop:annealed_posterior} states that, in the limit, the posterior expectation in (\ref{eq:latent_post_expectation}) simplifies  to (\ref{eq:asymptotic_expectation}) such that each particle maps to one DAG $\smash{\G_\infty(\Z^{(l)}_T)}$ and set of discrete targets $\smash{\I_\infty^\text{tar}(\intvGam_T^{(l)})}$.
After completing the SVGD steps, we hence return 
% $\{ (\G_\infty (\Z^{(l)}_T), \param^{(l)}_T, \I^{\text{tar}}_\infty (\intvGam^{(l)}_{T}),  \param^{(l)}_{\I, T})  \}_{l=1}^L$ 
the limit particles  $\smash{\G_\infty(\Z^{(l)}_T)}$ and $\smash{\I_\infty^\text{tar}(\intvGam_T^{(l)})}$ 
%together with $\param^{(l)}_T$ and $\param^{(l)}_{\I, T}$
as the particle approximation of the posterior
$p(\G, \param, \I_{\text{tar}}, \param_\I| \mathbf{D})$. The full procedure is summarized in Algorithm \ref{alg:bacadi_svgd} in Appx. \ref{appx:algo_details}.

%Overall, our framework allows us to perform gradient based approximate joint inference over discrete structures 
%Thus, the discrete particles $\G_\infty (\Z^{(l)}_T)$ and $\I_\infty$ are a consistent estimator of the initial expectation over $p(\G, \param, \I | \train)$.

The behavior of SVGD guarantees the minimization of the KL divergence and asymptotic convergence to the continuous posterior $p(\Z, \param, \intvGam, \param_{\I} | \mathbf{D})$ in the large sample limit of the number of particles \citep{liu2017stein}. By additionally annealing the continuous relaxations, the posterior that is approximated by SVGD converges to the semi-continuous posterior in Eq.~\ref{eq:posterior_unknown_interventions} we are ultimately interested in. 
Together, we thus obtain an asymptotically consistent approximation of $p(\G, \param, \I^{\text{tar}}, \param_{\I} | \mathbf{D})$.

\textbf{Identifiability.}\quad
Our approach focuses on joint Bayesian inference over CBNs and interventions and applies to many instantiations of the generative process in (\ref{eq:data_gen_process}). 
Under additional assumptions such as specific functional forms and noise distributions \citep{shimizu2006linear,hoyer2008nonlinear,peters2014identifiability}, identification results of related, non-Bayesian methods \citep[e.g.,][]{brouillard2020differentiable} apply to BaCaDI in the large sample limit where the posterior is dominated by the likelihood.
By proposing an inference technique rather than a specific parametric model, theoretical results on identification are not of direct concern to us, similar to related algorithmic works on structure learning
\citep{zheng2018dags,yu2019dag,lorch2021dibs}.

% In the large sample limit of data samples, the weight of the prior vanishes and the posterior is dominated by the likelihood, where identification results of related, non-Bayesian methods can apply to BaCaDI, e.g. by \citep{brouillard2020differentiable}. However, our approach is generally agnostic to the specifics of the generative process and does not make additional assumptions such as faithfulness which are necessary to provide identifiability results \citep{peters2016causal}.

Figure \ref{fig:toy_example} illustrates an example of the returned posterior particles for $\G$ and $ \I^{\text{tar}}$ alongside the ground truth for data generated by a linear-Gaussian, five-node CBN.
While SVGD yields a set of particles of with equal weights, we weight each particle by its unnormalized posterior probability $p(\G, \param, \param_{\I}, \I^{\text{tar}}, | \mathbf{D})$ for performing approximate Bayesian model averaging, similar to \citet{friedman2013data}. We find that this improves the empirical performance.

\vspacecaption
\section{EXPERIMENTS}
\vspacecaptionlow
\label{sec:eval}
\looseness-1
We evaluate \ours on different causal discovery tasks with data from multiple contexts. Our goal is to empirically investigate how accurately \ours predicts the causal structure as well as the intervention targets and how it compares to related state-of-the-art methods. First, we focus on synthetic datasets generated by CBNs. 
Second, we use the SERGIO simulator \citep{dibaeinia2020SERGIO} to evaluate the methods on simulated gene expression data. 
In this more realistic setting, the methods operate under significant model misspecification, since the data-generating processes do not match the priors and likelihood. % Finally, we analyze how well the uncertainty estimates are calibrated.
% \begin{figure}[t]
%     \centering
%     % \includegraphics[width=\linewidth]{figures/figure_1.pdf}
%     \includegraphics[width=\linewidth]{figures/toy_posterior.pdf}
%     \vspace{-20pt}
%     \caption{Visual comparison of samples from \ours's inferred posteriors $p(\G | \mathbf{D})$ and $p(\I^{\text{tar}} | \mathbf{D})$ with corresponding ground truths $\G_{\text{gt}}$ and $\I^{\text{tar}}_{\text{gt}}$. Blue colors represent presence of edges or targets. Here, we use a SF-2 graph with $d=5$ nodes and $M=3$ contexts.}
%     \label{fig:toy_example}
% \end{figure}
\begin{figure*}[t]
    \centering
    \vspacefigabove
    \begin{subfigure}{.49\textwidth}
        \centering
        \includegraphics[width=\linewidth]{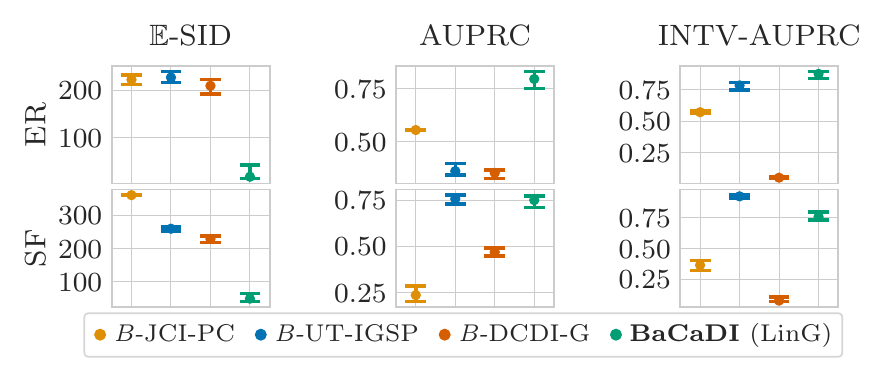}
        \vspacefigcaptionmid
        \vspace{-4pt}
        \caption{Linear Gaussian}
        % \includegraphics[width=\linewidth]{figures/prelim_lingauss_v2.pdf} 
        % \caption{\looseness -1 Joint posterior inference over CBNs and interventions for {\em linear Gaussian} ground-truth CBNs. The results are for data from ER-2 and SF-2 graphs with $d=20$ variables and $M = 20$ contexts.}
        % \vspacefigcaptionmid
        \label{fig:lingauss}
    \end{subfigure}\hfill
    \begin{subfigure}{.49\textwidth}
        \centering
        \includegraphics[width=1.\linewidth]{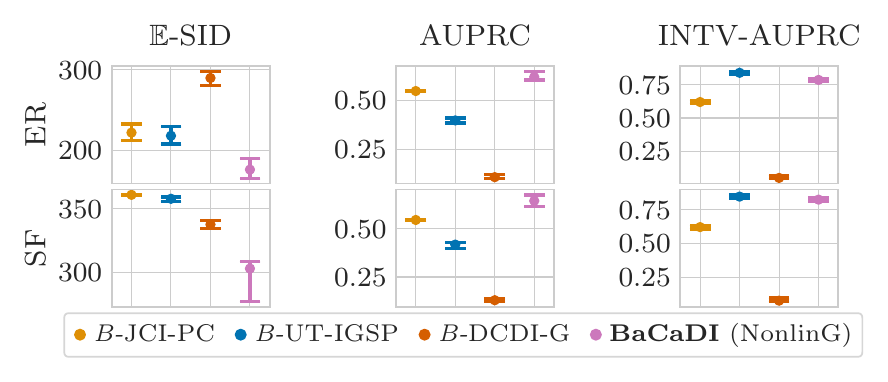}
        \vspacefigcaptionmid
        \vspace{-4pt}
        \caption{Nonlinear Gaussian}
        \label{fig:fcgauss}
    \end{subfigure}
    \vspacefigcaptionlow
    \caption{\looseness -1 Joint posterior inference over CBNs and interventions for {\em linear} and {\em nonlinear Gaussian} CBNs. The data is from ER-2 (top) and SF-2 (bottom) graphs with $d=20$ and $M = 20$ contexts. \ours consistently gives the best causal structure and intervention predictions. 
    % For $\E$-SID (AUPRC, INTV-AUPRC), lower (higher) values are better.
    Lower values are better for $\E$-SID. Higher values are better for AUPRC/INTV-AUPRC.
    % %For each metric, we report the median and the 90\% confidence interval.
    \vspacefigcaptionlow
    }
    %Joint posterior inference of graphs, parameters, intervention targets and intervention effects of {\em linear Gaussian CBNs} for ER-2 and SF-2 graphs with $d = 20$ nodes and $M = 20$ contexts. 
\end{figure*}
\vspacesubcaption
\subsection{Experimental Setup}
\vspacesubcaptionlow

\textbf{Datasets. }
\looseness -1 Following related work \citep{zheng2018dags, yu2019dag, zheng2020learning, annadani2021variational, scherrer2021learning, lorch2021dibs}, we perform inference of randomly sampled graphs. We consider \erdosrenyi~ \citep{gilbert1959random} and scale-free \citep{barabasi1999emergence} random graphs with $d=20$ nodes and $2d$ edges in expectation (ER-2 and SF-2).
We create datasets by randomly sampling CBN parameters or simulating data with SERGIO. We then split the data into a dataset that is used to perform inference and held-out test datasets which are used to compute metrics.
In all settings, we collect $n_0=100$ observational samples together with $n_k=10$ samples per intervention context for $k \in \{1,\dots,M\}$. This constitutes a low-sample setting commonly found in many applications. Further details on data generation are given in Appx.~\ref{appx:exp_details}.
%and we have fewer samples per context than actual variables of interest. 
%For detailed descriptions of data generation, we refer to Appx. ~\ref{appx:exp_details}.

\textbf{Baselines. }
\looseness-1
Since \ours is a probabilistic method, we compare it to {\em bootstrapped} versions of existing algorithms that are capable of joint causal inference from multiple contexts with unknown interventions. 
We benchmark \ours with constraint-based methods UT-IGSP \citep{squires2020permutationbased} and the JCI framework with the PC algorithm (JCI-PC) \citep{mooij2016multiple} and the score-based method DCDI \citep{brouillard2020differentiable}, which can handle unknown interventions. JCI-PC and UT-IGSP are based on conditional independence or invariance tests. For DCDI, we use a neural network to model the local conditionals with Gaussian additive noise (DCDI-G). 
% This is the same model class and capacity as \ours with a nonlinear likelihood model. % with one hidden layer with 5 logits.
%For our data that follows a Gaussian distribution, we thus
Since all of these methods infer only a single DAG estimate, we use the nonparametric DAG bootstrap approach \citep{friedman2013data, agrawal2019abcdstrategy} to obtain an approximate distribution over DAGs and intervention targets. 
We report the baselines with a \textit{``B-"} prefix to highlight that they are bootstrapped. Throughout the experiments, we use 20 bootstrap samples for all methods.

\textbf{\ours  instantiations. } We instantiate  \ours using 20 particles for SVGD, which matches the number of bootstrap samples of the baselines, and run it for 2000 steps of SVGD updates. In case a returned particle represents a cyclic graph, we drop the particle for the posterior approximation. Unless specified otherwise, we model interventions using a Gaussian likelihood $p_i^{I_k}(\evx_i| \param_{I_k}) = \mathcal{N}(x_i|\mu^I_{k,i}, \sigma_I)$ with fixed variance ${\sigma_I}^2=0.5$. We infer the means $\param_{I_k} = [\mu^I_{k,1}, ..., \mu^I_{k,d}]$ using a wide Gaussian prior $p(\param_{I_k} | \Itar_k) = \prod_{i \in \Itar_k} p(\mu^I_{k,i}| I^{\text{tar}}_{k,i}=1) = \prod_{i \in \Itar_k} \mathcal{N}(\mu^I_{k,i}|0, 10)$. This reflects an uninformative prior over a large effect range of the interventions. The conditionals for the observational likelihood are either modelled as linear or nonlinear models with additive Gaussian noise. The latter corresponds to the model of DCDI-G and uses 1 hidden-layer neural networks (NN) with $5$ hidden units. More details about the generative models can be found in Appx.~\ref{appx:model_details}.
% As explained in Sect.~\ref{sec:joint_bayesian_inference_cbns_interv}, the Bayesian formulation of \ours  allows us to incorporate application specific prior knowledge.

\textbf{Metrics. }
Our reported metrics focus on three essential aspects of our inference problem: causal discovery, intervention detection, and inference of the full CBN for modeling the effects of interventions. \vspacecaptionlow
\begin{itemize}[wide=0pt, leftmargin=8pt]
\item {\bf Causal structure:} The {\em Structural Interventional Distance} (SID) \citep{peters2015structural} quantifies the agreement between $\G$ and the true $\G_{\text{gt}}$ by the degree to which their adjustment sets coincide. Since we perform posterior inference, we report the \emph{expected} SID:
$\mathbb{E}\text{-SID}(p, \G_{\text{gt}}) := \sum_{\G} p(\G|\D) \cdot \text{SID}(\G, \G_{\text{gt}})$.
UT-IGSP and JCI-PC only return a CPDAGs of the Interventional Markov Equivalence Class (I-MEC), so we calculate its lower and upper bound SID and use their midpoint in the $\mathbb{E}$-SID.
We also report the area under the precision-recall curve (AUPRC) for individual edge predictions based on the posterior marginals $p(g_{ij}=1| \mathbf{D})$.
%Note that this is a more suitable metric than, e.g., the AUROC, as sparse graphs translate to highly imbalanced edge classification tasks.

\item {\bf Intervention targets}: We report {\em interventional AUPRC} (INTV-AUPRC) for the detection of the individual target variables in each environment.
%capturing how well an algorithm predicts which variables have been intervened on.

\item {\bf Intervention effects:} We report the average negative {\em interventional log-likelihood} (I-NLL) on $M^{\text{test}} = 10$ heldout interventional datasets $\mathbf{D}^{\text{test}} = \{\train^{\text{test}}_1, ..., \train^{\text{test}}_{10} \}$, where different interventions are performed than in the training datasets. Each test dataset comprises 100 samples and has known intervention targets $\Itar_{\text{test}, k}$ and effect distributions $p(x_i|\param_{I_{\text{test}, k}})$.
% \begin{align*}
%     \text{I-LL}(p, \mathbf{D}^{\text{test}}) := -  \frac{1}{M^{\text{test}}} \sum_{k=1}^{M^{\text{test}}}  \E_{p(\G, \param | \mathbf{D})} \left[ \frac{1}{|\train^{\text{test}}_k|} \log p(\train^{\text{test}}_k | \G,\param,\Itar_{\text{test}, k}, \param_{I_{\text{test}, k}}) \right] \;.
% \end{align*}
Since UT-IGSP and JCI-PC do not learn the conditional distributions, we use the closed-form MLE parameters for a linear Gaussian model to compute the heldout I-NLL \citep{Hauser_2014}.
\end{itemize}

\vspace*{-7pt}
Additional details on the metrics are given in Appendix ~\ref{appx:experiment_metrics}.

\textbf{Result aggregation. }
For all methods and settings, we perform a search over a specified range of hyperparameter using at least $20$ settings. For the specific choice of hyperparameters, we collect results over 30 different random task instances and pick the hyperparameters that resulted in the lowest I-NLL across the 30 instances in the held-out interventional dataset. We report the median of each metric together with its 90\% confidence interval based on the empirical percentiles. 
%All methods are used with $20$ posterior samples (bootstrap samples) apart from DCDI. \ours  is run for $3000$ iterations. 

\vspacesubcaption
\subsection{Experiment Results on Synthetic Data}
\vspacesubcaptionlow
\label{sec:synthetic_exp_results}

\looseness-1
% First, we study the performance of \ours and the baselines on joint causal inference from synthetic data. 
In the synthetic data setting, we focus on hard interventions that set the values of the targeted variable to samples from a Gaussian with a randomly chosen mean bounded away from zero and a variance of $0.5$.
%Our focus is on hard interventions that cut the dependencies to the parents and set the variable to a Gaussian with a randomly chosen mean bounded away from zero (i.e., not fixed a-priori) and small noise. 
Each interventional dataset is generated by intervening on a specific variable. We target all variables in the graph, i.e., the number of interventional contexts is $M=d$. In total, this results in $300$ samples for synthetic datasets with $d=20$ variables and $n_0=100, n_k=10$.

We first consider {\em linear Gaussian} CBNs, where each variable is a linear combination of its parents with additive Gaussian noise. The corresponding results for ER-2 and SF-2 are shown in Fig. \ref{fig:lingauss}. As a second task, we evaluate the setting of {\em nonlinear Gaussian} conditionals, where the mean of each variable given its parents is described by a neural network (NN) (see Appx. \ref{appx:data_generation} for details). Fig. \ref{fig:fcgauss} displays the results for nonlinear CBNs.

Across these four synthetic evaluation settings, \ours's causal structure predictions are the closest to the ground-truth CBN in terms of their intervention implications ($\mathbb{E}$-SID) and the prediction of individual edges (AUPRC). In most cases, our method outperforms the baselines by a significant margin. Moreover, \ours achieves strong AUPRC scores for the prediction of intervention targets. Among the baselines, UT-IGSP is the best at detecting interventions, largely on par with \ours. Although UT-IGSP also achieves high AUPRC for the linear SF-2 setting, the method performs significantly worse than \ours in other settings and metrics.  
DCDI perfoms poorly in predicting the interventions and causal mechanisms. 
% We hypothesize that this is because DCDI neither takes into account epistemic uncertainties nor regularizes its neural network conditionals, making it prone to overfitting in small data settings like ours.

% We hypothesize that bootstrapping performs so much worse because, 
In contrast to our fully Bayesian treatment of the joint posterior $p(\G, \param, \I^\text{tar}|\mathbf{D})$, the bootstrap baselines cannot propagate epistemic uncertainty between the inference of intervention targets and effects and inference of the CBN. This makes methods like DCDI, which does not regularize its neural network conditionals, prone to overfitting in small data settings, and may explain why \ours is the only method to perform reliably in the low-sample regime. 
% \TODO{possible some concluding remarks here? hypotheses why methods perform worse? low samples?}  
% \TODO{talk about DCDI one we have an explanation why it predicts interventions everywhere}
% \begin{figure*}[t]
%     \centering
%     % \includegraphics[width=\linewidth]{figures/prelim_fcgauss_v2.pdf}
%     \vspacefigabove
%     \includegraphics[width=.6\linewidth]{figures/fcgauss_metrics.pdf}
%     \vspacefigcaptionlow
%     \caption{\looseness -1 Joint posterior inference over CBNs and interventions for {\em nonlinear Gaussian} ground-truth CBNs. The results are for data from ER-2 (top) and SF-2 (bottom) graphs with $d=20$ variables and $M = 20$ contexts. \ours consistently gives the best causal structure and intervention predictions. 
%     %For each metric, we report the median and the 90\% confidence interval.
%     \vspacefigcaptionmid
%     }
%     \label{fig:fcgauss}
% \end{figure*}

\textbf{Additional analyses.} We perform additional experiments %to investigate the scalability of \ours 
for $d=50$ node graphs as well as larger datasets. The results are shown in Fig.\ref{fig:lingauss_moredata}-\ref{fig:fcgauss_moredata} and \ref{fig:lingauss_50node} in Appx. \ref{appx:additional_experiments}.  In terms of scaling to larger graphs, \ours is competitive to current state-of-the-art methods of causal discovery. For substantially larger graphs, the computational cost become prohibitively high when basing inference on SVGD. As an interesting use case, Appx. \ref{appx_sec:obs_vs_intv} contains an ablation study investigating how \ours can leverage interventional data compared to observational data, even when targets are unknown.

\vspacecaption
\subsection{Experiments with Gene-Regulatory Networks}
\vspacesubcaptionlow
\label{sec:sergio_exps}

\looseness -1 We evaluate all methods in a realistic application domain using SERGIO \citep{dibaeinia2020SERGIO}, a simulator for single-cell expression data of gene regulatory networks. Given a user-defined causal graph $\G$, SERGIO utilizes stochastic differential equations to simulate the gene expression dynamics and generate realistic single-cell transcriptomic datasets, which correspond to samples from the steady state of this dynamical system. Since real-world gene regulatory networks resemble scale-free structures \citep{albert2005scale, ouma2018topological}, we use randomly sampled SF-2 graphs with $d=20$. 
In this domain, we perform $M=10$ gene knockout interventions on a single randomly selected target per context, resulting in a dataset size of $200$ including the observational samples. 
The data is standardized before inference. 
See Appx. \ref{appx:sergio_details} for more details. 

% SERGIO then simulates every gene's expression dynamics using a stochastic differential equation called the chemical Langevin equation.
\looseness -1 The Bayesian formulation allows \ours to embed prior knowledge into the inference process in a principled manner. Since we perform knockout interventions, we expect intervention values to be close to zero and exhibit low variance. To reflect this belief, we set the intervention noise to ${\sigma_I}^2=0.01$ and use the prior $p(\mu^I_{k,i}| I^{\text{tar}}_{k,i}=1) = \mathcal{N}(\mu^I_{k,i}|0, 1)$.
%to reflect our prior belief that the intervention effects are close to $0$.
%Since, \ours is a fully Bayesian model, we can seamlessly embed prior knowledge about our problem at hand into the inference process. 

\begin{figure}
    \centering
    \vspacefigabove
    \includegraphics[width=\linewidth]{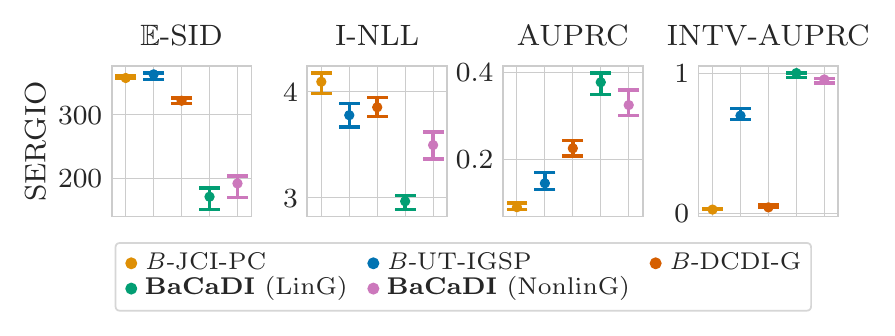}
    \vspacefigcaption
    \caption{
    \looseness-1
    %  Joint posterior inference over CBNs and interventions for {\em nonlinear Gaussian} ground-truth CBNs. The results are for data from ER-2 (top) and SF-2 (bottom) graphs with $d=20$ nodes and $M = 20$ contexts. \ours consistently gives the best causal structure and intervention predictions.
    Joint posterior over CBNs and interventions for simulated {\em gene expression} data with $d = 20$ and $M = 10$ contexts. \ours makes significantly better causal mechanism predictions than the baselines and accurately identifies the intervention targets. 
    % For $\E$-SID and I-NLL (AUPRC and INTV-AUPRC), lower (higher) values are better.
    Lower values are better for $\E$-SID/I-NLL. Higher values are better for AUPRC/INTV-AUPRC.
    \vspacefigcaptionmid}
    \label{fig:SERGIO}
\end{figure}
\textbf{Results. }
Fig. \ref{fig:SERGIO} shows the results for the SERGIO datasets. As in the synthetic CBN domain, \ours  infers the ground-truth graph most accurately. % given the provided data. 
Moreover, our method is very accurate in predicting intervention targets as reflected by the intervention target AUPRC, where it benefits from the prior. In this context, the accuracy of JCI-PC is close to random guessing, predicting nearly no interventions and edges. % in the CPDAG. 
%This is mostly likely because the conditional dependence tests fail to reach significance given the small amount of available data.

\looseness -1 Since the data generated by SERGIO are samples from the steady state of a stochastic dynamical system, our Bayesian model from Sec. \ref{sec:diff_model_over_cbs_interv} is misspecified. The fact that \ours still performs well demonstrates its robustness to model mismatch in practice.
Overall, \ours is able to make accurate causal structure predictions with only 200 simulated gene expression measurements from a challenging multi-experiment setting. This is a promising step towards joint causal inference in the life sciences. Here, the predicted intervention targets can further be of independent interest, e.g., for understanding off-target effects of drugs. 

\looseness-1
We believe that evaluating causal discovery methods beyond synthetic toy datasets on more realistic benchmarks, such as SERGIO, is an important direction for future research. In particular, there are many open questions about the practical applicability of current algorithms to real-world data. We refer to the work of \citet{reisach2021beware} who discuss, in detail, potential issues that arise from synthetic benchmarking and the scale sensitivity of continuous optimization for causal structure learning.
\vspacecaption
\section{DISCUSSION}
\vspacecaption
\label{sec:conclusion}
In this work, we introduced \ours , a fully-differentiable Bayesian causal discovery framework for data generated under various unknown interventional conditions. \ours  performs approximate inference jointly over the underlying causal graph, mechanisms, and the unknown interventions from multiple contexts. A key feature of \ours is its principled end-to-end treatment of epistemic uncertainty.
In our experiments, the naive bootstrapping of previous methods performs worse, potentially because the epistemic uncertainty does not propagate between the unknown interventions and the unknown causal Bayesian network.
\ours, on the other hand, operates reliably even when data is scarce, and can be instantiated with any parametric model as well as specific prior knowledge from domain experts.
% providing well-calibrated uncertainty estimates alongside its structural predictions.

\looseness-1
While the Bayesian approach has shown promise in causal discovery, there are interesting open problems that remain. One is that the posterior becomes strongly intractable for graphs larger than five nodes, making it difficult to accurately characterize or quantify the quality of the posterior approximation. Additionally, since it is driven by the likelihood term, the posterior clusters data points based on their match to the observational likelihood or whether they require a separate likelihood. As a result, the performance of identifying interventions is improved when there is a stronger shift in distribution; interventions that only slightly alter the distribution cannot be detected with limited data.
% interesting open question that remains for Bayesian methods for causal discovery is characterizing the true posterior and its properties such as sparsity; since 

\looseness-1 Our work is motivated by the challenging problem of inferring the causal mechanisms of gene regulatory networks from real single-cell gene expression data. Our experimental results for the simulated gene expression data in Sec. \ref{sec:sergio_exps} show that \ours provides an important step towards achieving this goal. To ultimately reach it, future work needs to address a range of further challenges, such as dealing with the experimental measurement noise incurred by single-cell sequencing techniques.

% Given these positive results, there are promising prospects of applying \ours to causal discovery in fields like molecular biology, as it pushes the boundary of what causal inferences we can draw from experimental data in the life sciences.\bernhard{last sentence would better fit into the conclusions, I suggest to move it} 

%Future: extend with soft interventions (more realistic) e.g. with new interventional BGe score \citep{kuipers2022iBGE} 
\acknowledgments{

% \subsubsection*{Acknowledgements}

This research was supported by the European Research Council (ERC) under the European Union's Horizon 2020 research and innovation program grant agreement no.\ 815943 and the Swiss National Science Foundation under NCCR Automation, grant agreement 51NF40 180545.
Jonas Rothfuss was supported by an Apple Scholars in AI/ML fellowship. We thank Parnian Kassraie for valuable feedback.
}
% \newpage
\bibliography{references}

\clearpage

\onecolumn
\appendix
%   \hrule height 4 pt
%   \vskip 0.25in
%   \vskip -\parskip%
% {\centering
% {\LARGE\bf Supplementary Material:\\ BaCaDI: Bayesian Causal Discovery \\ with Unknown Interventions \par }}
%   \vskip 0.29in
%   \vskip -\parskip
%   \hrule height 1 pt
%   \vskip 0.09in%

\section{PROOFS \& DERIVATIONS}
\label{appx:proofs}

\subsection{Latent Posterior (Proof of Proposition \ref{prop:latent_post_expectation})}
\label{appx:proof_latent_posterior_expectation}
Recall that we have the generative model under which we factorize
\begin{align*}
    p(\Z, \G, \param, \intvGam, \I, \mathbf{D}) & = \underbrace{p(\Z)p(\G|\Z) p(\param|\G)}_{\text{gen. process CBN}} \prod_{k=1}^{M} \underbrace{p(\intvGam_k)p(\Itar_k|\intvGam_k) p(\param_{I_k} | \Itar_k)}_{\text{gen. process interv.}} \underbrace{p(\mathcal{D}_k|\G, \param, \Itar_k, \param_{I_k})}_{\text{interv. likelihood}} \vspace{-10pt}
\end{align*}
where we write $\intvGam := [\intvGam_1,...,\intvGam_M]$. For brevity, we can express this as
\begin{align*}
    p(\Z, \G, \param, \intvGam, \I, \mathbf{D}) & = p(\Z)p(\G|\Z) p(\param|\G) \cdot p(\intvGam) p(\I^{\text{tar}} | \intvGam) p(\param_\I|\I^{\text{tar}}) \cdot p(\mathbf{D} | \G, \I, \param) \vspace{-10pt}
\end{align*}
where $p(\mathbf{D} | \G, \I, \param) = p(\mathbf{D} | \G, \param, \I^{\text{tar}}, \param_\I)=\prod_{k=1}^{M} p(\mathcal{D}_k|\G, \param, \Itar_k, \param_{I_k})$.

This gives us
This gives us
\begin{align*}
     &  \E_{p(\G, \param, \I | \mathbf{D})} [f(\G, \param, \I)] \\
     & = \sum_{\G} \int_\param \sum_{\I^{\text{tar}}} \int_{\param_\I} p(\G, \param, \I | \mathbf{D}) f(\G, \param, \I) d\param d\param_\I  \\
     & = \sum_{\G} \int_\param \sum_{\I^{\text{tar}}} \int_{\param_\I} p(\G, \I^{\text{tar}}, \param, \param_\I| \mathbf{D}) f(\G, \param, \I) d\param d\param_\I  \\
     & (\text{splitting } \I \text{ to } \I^{\text{tar}}, \param_\I)\\
     & = \sum_{\G} \int_\param \sum_{\I^{\text{tar}}} \int_{\param_\I} \frac{p(\G, \I^{\text{tar}}, \param, \param_\I, \mathbf{D}) f(\G, \param, \I)}{p(\mathbf{D})} d\param d\param_\I  \\
     & = \sum_{\G} \int_\param \sum_{\I^{\text{tar}}} \int_{\param_\I} \int_\Z \int_\intvGam \frac{p(\Z, \G, \intvGam, \I^{\text{tar}}, \param, \param_\I, \mathbf{D}) f(\G, \param, \I)}{p(\mathbf{D})} d\Z d\intvGam d\param d\param_\I  \\
     & (\text{extending by } \Z, \intvGam )\\
     %& = \text{by the generative model})\\
     & = \int_{\Z, \intvGam, \param, \param_\I} \sum_{\G, \I^{\text{tar}}}  \frac{p(\Z, \G, \intvGam, \I^{\text{tar}}, \param, \param_\I, \mathbf{D}) f(\G, \param, \I)}{p(\mathbf{D})} d\Z d\intvGam d\param d\param_\I  \\
     & (\text{rearranging})\\
     & = \int_{\Z, \intvGam, \param, {\param_\I}}  \sum_{\G, \I^{\text{tar}}}  \frac{p(\Z)p(\G|\Z) p(\param|\G) p(\intvGam) p(\I^{\text{tar}} | \intvGam) p(\param_\I|\I^{\text{tar}}) p(\mathbf{D} | \G, \I, \param) f(\G, \param, \I)}{p(\mathbf{D})} d\Z d\intvGam d\param d\param_\I  \\
     & (\text{by the generative model})\\
     & = E_{p(\Z, \param, \intvGam, \param_\I |  \mathbf{D})} \left[ \sum_{\G, \I^{\text{tar}}} \frac{p(\G|\Z) p(\param|\G) p(\I^{\text{tar}} | \intvGam) p(\param_\I|\I^{\text{tar}}) p(\mathbf{D} | \G, \I, \param) f(\G, \param, \I)}{p(\param, \param_\I, \mathbf{D} | \Z, \intvGam)} \right] \\
     & (\text{since } p(\Z, \param, \intvGam, \param_\I |  \mathbf{D}) = \frac{p(\Z) p(\intvGam) p(\param, \param_\I, \mathbf{D} | \Z, \intvGam) }{p(\mathbf{D})} )\\
     % & = \text{by the generative model})\\
     & = E_{p(\Z, \param, \intvGam, \param_\I |  \mathbf{D})} \left[  \frac{\sum_{\G, \I^{\text{tar}}} p(\G|\Z) p(\param|\G) p(\I^{\text{tar}} | \intvGam) p(\param_\I|\I^{\text{tar}}) p(\mathbf{D} | \G, \I, \param) f(\G, \param, \I)}{p(\param, \param_\I, \mathbf{D} | \Z, \intvGam)}  \right]  \\
     & (\text{rearranging )}\\
     %&  \text{by the generative model})\\
     & = E_{p(\Z, \param, \intvGam, \param_\I |  \mathbf{D})} \left[ \frac{\sum_{\G, \I^{\text{tar}}} p(\G|\Z) p(\param|\G) p(\I^{\text{tar}} | \intvGam) p(\param_\I|\I^{\text{tar}}) p(\mathbf{D} | \G, \I, \param) f(\G, \param, \I)}{\sum_{\G, \I^{\text{tar}}} p(\G, \I^\text{tar}, \param, \param_\I, \mathbf{D} | \Z, \intvGam)} \right]  \\
     & (\text{law of total probability )}\\
     & = \E_{p(\Z, \param, \intvGam, \param_\I |  \mathbf{D})} \left[ \frac{\sum_{\G, \I^{\text{tar}}} p(\G|\Z) p(\param|\G) p(\I^{\text{tar}} | \intvGam) p(\param_\I|\I^{\text{tar}}) p(\mathbf{D} | \G, \I, \param) f(\G, \param, \I)}{\sum_{\G, \I^{\text{tar}}} p(\G|\Z) p(\param|\G) p(\I^{\text{tar}} | \intvGam) p(\param_\I|\I^{\text{tar}}) p(\mathbf{D} | \G, \I, \param)} \right]\\
     & (\text{by the generative model )}\\
     & = \E_{p(\Z, \param, \intvGam, \param_\I |  \mathbf{D})} \left[\frac{\E_{p(\G|\Z),p(\I^{\text{tar}}|\intvGam)}[f(\G, \param, \I)p(\param|\G)p(\param_\I|\I^{\text{tar}})p(\mathbf{D}|\G,\I,\param)]}{\E_{p(\G|\Z),p(\I^{\text{tar}}|\intvGam)}[p(\param|\G)p(\param_\I|\I^{\text{tar}})p(\mathbf{D}|\G,\I,\param)]} \right] \\
     \label{appx:latent_post_expectation_proof}
\end{align*}

\begin{flushright}$\square$ \end{flushright}

\subsection{Annealing the Posterior (Proof of Proposition \ref{prop:annealed_posterior})}

\label{appx:annealing}
The latent variables $\Z$ and $\intvGam$ probabilistically model the causal graph $\G$ and intervention target masks $\I^\text{tar}$. In a similar way, they can be viewed as as continuous relaxations of $\G$ and $\I^\text{tar}$ respectively, where the $\alpha$ trades-off between smoothness and accuracy of these relaxations. If $\alpha \rightarrow \infty$, the sigmoid $\sigma_\alpha(\cdot)$ converges to the unit step function so that the probability distributions 
\begin{align}
p(\G_{i,j}|\Z) &= p(\G_{i,j}|\mathbf{u}_i, \mathbf{v}_j) = \mathbf{1}_{(\mathbf{u}_i^\top \mathbf{v}_j > 0) = \G_{i,j}} ~, \text{and} \\
p(\Itar_{k,i}  | \intvGam) &= p(\Itar_{k,i} | \gamma_{k, i}) = \mathbf{1}_{(\gamma_{i, k} > 0) = \Itar_{k,i}}
\end{align}
become deterministic indicator functions, representing only single discrete graph $\G_\infty(\Z) \in \{0,1\}^{d \times d}$ and target mask $\I_\infty^\text{tar}(\intvGam) \in \{0,1\}^{M \times d}$. Since, the probability distributions $p(\G|\Z)$ and $p(\I^{\text{tar}}|\intvGam)$ converge to indicator function, 
%imply 1-1 mappings between $\G$ and $\Z$ as well as $\I^{\text{tar}}$ and $\intvGam$ 
they allow us to simplify the expectation in Proposition \ref{prop:latent_post_expectation} as follows:
% \begin{align}
%     & p(\G, \param, \I | \mathbf{D}) \\
%     & = \iint p(\Z, \param, \intvGam, \param_\I |  \mathbf{D}) \frac{p(\G|\Z) p(\I^{\text{tar}} | \intvGam) p(\param_\I|\I^{\text{tar}}) p(\param, \mathbf{D} | \G, \I, \param) } {\E_{p(\G^\prime|\Z),p(\I^{\text{tar}\prime}|\intvGam)}[ p(\param_\I|\I^{\text{tar}\prime}) p(\param, \mathbf{D} | \G^\prime, \I^\prime, \param)]} d\Z d\intvGam \\
%     & \rightarrow  \iint p(\Z, \param, \intvGam, \param_\I |  \mathbf{D}) p(\G|\Z) p(\I^{\text{tar}} | \intvGam) d\Z d\intvGam \\
%     & =  p(\G(\Z_\infty), \param, I_\infty^\text{tar}(\intvGam), \param_\I |  \mathbf{D})\\
%     & = \iint p(\Z, \param, \intvGam, \param_\I |  \mathbf{D}) \mathbf{1}_{\G =  \G_\infty(\Z)} \mathbf{1}_{I_\infty^\text{tar}(\intvGam)} d\Z d\intvGam 
% \end{align}

\begin{equation}
\begin{split}
     &  \E_{p(\G, \param, \I, \param_\I  | \mathbf{D})} [f(\G, \param, \I, \param_\I )] \\
     & = \E_{p(\Z, \param, \intvGam, \param_\I |  \mathbf{D})} \left[\frac{\E_{p(\G|\Z),p(\I^{\text{tar}}|\intvGam)}[f(\G, \param, \I, \param_\I )p(\param|\G)p(\param_\I|\I^{\text{tar}})p(\mathbf{D}|\G,\I,\param)]}{\E_{p(\G|\Z),p(\I^{\text{tar}}|\intvGam)}[p(\param|\G)p(\param_\I|\I^{\text{tar}})p(\mathbf{D}|\G,\I,\param)]} \right] \\ \\
     & \rightarrow \E_{p(\Z, \param, \intvGam, \param_\I |  \mathbf{D})} [f(\G_\infty(\Z), \param, \I_\infty^\text{tar}(\intvGam), \param_\I )]\;.
     \label{appx:latent_post_expectation_infty}
    \end{split}
\end{equation}
This holds since the inner expectations evaluate to a single point for $\alpha \rightarrow \infty$ and thus cancel out.

If additionally the inverse temperature parameter in the latent prior $p_\beta(\Z)$ goes to infinity, i.e., $\beta \rightarrow \infty$, $p_\beta(\Z)$, the support of $p_\beta(\Z) = p_\beta(\mathbf{U}, \mathbf{V})$ reduces to the set  $\text{supp}(p_\infty(\Z)) = \{\Z | \Z \in \R^{2 \times d \times d} \wedge  h(\G_\infty(\Z)) = 0\}$. In particular, the prior only assigns non-zero probabilities to latent variables $\Z$ that correspond to acyclical graphs $\G_\infty(\Z)$.
Since the posterior $p(\Z, \G, \param, \intvGam, \I^\text{tar},  \param_\I | \mathbf{D})$ depends multiplicatively on $p_\beta(\Z)$ (see Eq. \ref{eq:data_gen_process}), the support of 
\begin{equation}
    p(\G, \I^\text{tar} | \mathbf{D}) = \int_{\Z, \intvGam, \param, \param_\I} p(\Z, \G, \param, \intvGam, \I^\text{tar},  \param_\I | \mathbf{D}) d\Z d\intvGam d\param d\param_\I 
\end{equation}
asymptotically becomes a subset of $\{\mathbf{G} | \mathbf{G} \in \{0,1\}^{n \times n} \wedge \mathbf{G} \text{ is acyclic} \} \times \{0,1\}^{M \times d}$ since $h(\G) = 0 \Leftrightarrow \G \text{ is acyclic}$.

\subsection{Scores}
\label{appx:scores}
In this section, we show how to derive the score, i.e. the gradient of the log-posterior, of our model introduced in Sec. \ref{sec:diff_model_over_cbs_interv}.

The score w.r.t. auxiliary variables $\Z$ and $\intvGam$ requires marginalization over the corresponding discrete structures $\G$ and $\I^{\text{tar}}$ as given by
\begin{equation}
    \nabla_{\Z} \log p(\Z, \param, \intvGam, \param_{\I} | \mathbf{D}) = \nabla_{\Z} \log p(\Z) +\frac{\nabla_{\Z} \E_{p(\G | \Z)} \E_{p(\I^{\text{tar}} | \intvGam)}\left[   p(\param, \param_{\I}, \mathbf{D} | \G, \I^{\text{tar}})\right]}{\E_{p(\G | \Z)} \E_{p(\I^{\text{tar}} | \intvGam)}\left[   p(\param, \param_{\I}, \mathbf{D} | \G, \I^{\text{tar}})\right]} \;, \label{eq:score_Z}
\end{equation}
% \vspaceequation
\begin{equation}
    \nabla_{\intvGam} \log p(\Z, \param, \intvGam, \param_{\I} | \mathbf{D}) = \nabla_{\intvGam} \log p(\intvGam) +\frac{\nabla_{\intvGam} \E_{p(\G | \Z)} \E_{p(\I^{\text{tar}} | \intvGam)}\left[   p(\param, \param_{\I}, \mathbf{D} | \G, \I^{\text{tar}})\right]}{\E_{p(\G | \Z)} \E_{p(\I^{\text{tar}} | \intvGam)}\left[   p(\param, \param_{\I}, \mathbf{D} | \G, \I^{\text{tar}})\right]} \;. \label{eq:score_interv_gamma}
\end{equation} 
where, $p(\param, \param_{\I}, \mathbf{D} | \G, \I^{\text{tar}}) = p(\param|  \G) \prod_{k=1}^m p(\train_k | \G, \param, \Itar_k, \param_{I_k}) p(\param_{I_k}|\Itar_k)$.

The expectations in Eq. \ref{eq:score_Z} and Eq. \ref{eq:score_interv_gamma} can be estimated and differentiated by sampling the intervention masks $\I^{\text{tar}} \sim \text{Bern}\left(\sigma_\alpha(\intvGam)\right)$ and the adjacency matrices $\G \sim \text{Bern}\left(\sigma_\alpha(\mathbf{U}\mathbf{V}^\top)\right)$ using the Gumbel-Softmax trick \citep{jang2016categorical, maddison2017concrete}. For obtaining a differentiable log-likelihood, we mask individual log-likelihood summands based on $\G$ \citep{lorch2021dibs} and use the intervention masks $\mathcal{I}^{\text{tar}}$ to select between observational and interventional log-likelihood (cf. Eq.~\ref{eq:likelihood}).

To see how the above equations hold, we show the derivations in the following. We start the derivations with the unnormalized posterior since
\begin{align}
\nabla_{\Z} \log p(\Z, \intvGam, \param, \param_{\I} | \train) = & \nabla_{\Z} \log p(\Z, \intvGam, \param, \param_{\I}, \train) - \nabla_{\Z} \log p(\train) \\
= & \nabla_{\Z} \log p(\Z, \intvGam, \param, \param_{\I}, \train)
\end{align}
and analogously for the gradients w.r.t. to other variables $\Z, \intvGam, \param, \param_{\I}$ .
% \begin{align}
% \nabla_{\intvGam} \log p(\Z, \intvGam, \param | \train) = \nabla_{\intvGam} \log p(\Z, \intvGam, \param , \train) - \nabla_{\intvGam} \log p(\train) = \nabla_{\intvGam} \log p(\Z, \intvGam, \param , \train)
% \end{align}

By basic rules of probability theory and using the identity $\nabla_\rvx \log p(\rvx) = \nabla_\rvx p(\rvx)/ p(\rvx)$, we obtain
\begin{align}
    \nabla_{\Z}  \log p(\Z, \intvGam, \param, & \param_{\I}, \mathbf{D}) =   \\
    = & \nabla_{\Z} \log p(\Z) + \nabla_{\Z} \log p(\param, \param_{\I}, \mathbf{D} | \Z, \intvGam) \\
    = & \nabla_{\Z} \log p(\Z) +\frac{\nabla_{\Z} p(\param, \param_{\I}, \mathbf{D} | \Z, \intvGam)}{p(\param, \param_{\I}, \mathbf{D} | \Z, \intvGam)} \\
    = & \nabla_{\Z} \log p(\Z) +\frac{\nabla_{\Z} \left[ \sum_{\G} \sum_{\I^{\text{tar}}} p(\G | \Z) p(\I^{\text{tar}} | \intvGam) p(\param, \param_{\I}, \mathbf{D} | \G, \I^{\text{tar}})\right]}{\sum_{\G} \sum_{\I^{\text{tar}}} p(\G | \Z) p(\I^{\text{tar}} | \intvGam) p(\param, \param_{\I}, \mathbf{D} | \G, \I^{\text{tar}})} \\
    = & \nabla_{\Z} \log p(\Z) +\frac{\nabla_{\Z} \E_{p(\G | \Z)} \E_{p(\I^{\text{tar}} | \intvGam)}\left[   p(\param, \param_{\I}, \mathbf{D} | \G, \I^{\text{tar}})\right]}{\E_{p(\G | \Z)} \E_{p(\I^{\text{tar}} | \intvGam)}\left[   p(\param, \param_{\I}, \mathbf{D} | \G, \I^{\text{tar}})\right]} 
\end{align}
and analogously 
\begin{align}
        \nabla_{\intvGam}  \log p(\Z, \intvGam, \param, & \param_{\I}, \mathbf{D}) =   \\
    = & \nabla_{\intvGam} \log p(\intvGam) + \nabla_{\intvGam} \log p(\param, \param_{\I}, \mathbf{D} | \Z, \intvGam) \\
    = & \nabla_{\intvGam} \log p\intvGam) +\frac{\nabla_{\intvGam} p(\param, \param_{\I}, \mathbf{D} | \Z, \intvGam)}{p(\param, \param_{\I}, \mathbf{D} | \Z, \intvGam)} \\
    = & \nabla_{\intvGam} \log p(\intvGam) +\frac{\nabla_{\intvGam} \left[ \sum_{\G} \sum_{\I^{\text{tar}}} p(\G | \Z) p(\I^{\text{tar}} | \intvGam) p(\param, \param_{\I}, \mathbf{D} | \G, \I^{\text{tar}})\right]}{\sum_{\G} \sum_{\I^{\text{tar}}} p(\G | \Z) p(\I^{\text{tar}} | \intvGam) p(\param, \param_{\I}, \mathbf{D} | \G, \I^{\text{tar}})} \\
    = & \nabla_{\intvGam} \log p(\intvGam) +\frac{\nabla_{\intvGam} \E_{p(\G | \Z)} \E_{p(\I^{\text{tar}} | \intvGam)}\left[   p(\param, \param_{\I}, \mathbf{D} | \G, \I^{\text{tar}})\right]}{\E_{p(\G | \Z)} \E_{p(\I^{\text{tar}} | \intvGam)}\left[   p(\param, \param_{\I}, \mathbf{D} | \G, \I^{\text{tar}})\right]} \;. 
\end{align}

\section{ALGORITHM DETAILS}
\label{appx:algo_details}

\subsection{Annealing Schedule}
As described in Sec. \ref{appx:annealing}, we anneal $\alpha_t$ such that $\alpha_t \rightarrow \infty$ and, at terminal iteration of the SVGD training loop, convert $\Z$ and $\intvGam$ into their discrete counterparts $\G_\infty(\Z)$ and $\I_\infty^\text{tar}(\intvGam)$. Similarly, we set an annealing schedule $\beta_t \rightarrow \infty$ for inverse temperature parameter in the latent prior $p_\beta(\Z)$ such that we only model DAGs as the training progresses.

The annealing schedule for both $\alpha_t$ and $\beta_t$is a simple linear schedule, i.e. we put $\alpha_t = t \cdot \alpha$ and $\beta_t = t \cdot \beta$ for scale values $\alpha, \beta > 0$ where $t$ is the step in the SVGD inference (see \ref{appx:algo_details}). For all results in the main text, we fix $\alpha=0.01$ and $\beta=2$. These values were empirically found to be the best trade-off between smoothness and accuracy of the relaxations during the inference process.

\subsection{SVGD Kernel}
\label{appx:svgd_kernel}

In our experiments, we employ an additive RBF kernel. We also considered a product of RBF kernels, though, found that the additive kernel composition performed better. This additive kernel is defined as
\begin{align}
\begin{split}
    k \left( (\Z, \param, \intvGam, \param_{\I}), (\Z', \param', \intvGam', \param_{\I}') \right) := & \exp \left( - \frac{\lVert \Z - \Z' \rVert^2}{2 \tau_{Z}}\right) + \exp \left( - \frac{\lVert \intvGam - \intvGam' \rVert^2}{2 \tau_{\gamma}}\right) \\
    +&  \exp \left( - \frac{\lVert \param - \param' \rVert^2}{2 \tau_{\theta}}\right) + \exp \left( - \frac{\lVert \param_{\I} - \param_{\I}' \rVert^2}{2 \tau_{\theta}}\right) \label{eq:kernel}
\end{split}
\end{align}
with lengthscales $\tau_{Z}, \tau_{\gamma}, \tau_{\theta}$. For brevity, we also write $k(\mathbf{\Omega}, \mathbf{\Omega}')$ for (\ref{eq:kernel}) where $\mathbf{\Omega} := (\Z, \param, \intvGam, \param_{\I})$. For SVGD, the kernel introduces repulsive forces which make the particles disperse well across the domain (see \ref{appx:svgd}). 
%Importantly, the lengthscale hyperparameters provide the possibility to fine-tune the repulsion and hence calibrate our inference model. We show this in Sec. \ref{appx:experiments_calibration}.

\subsection{Algorithm overview}
We provide a pseudocode of our algorithm with its SVGD instantiation in Alg. \ref{alg:bacadi_svgd}.
\begin{algorithm}[t]
  \renewcommand{\algorithmicrequire}{\textbf{Input:}}
  \caption{BaCaDI with SVGD for inference of $p(\G, \param, \I^{\text{tar}}, \param_{\I} | \mathbf{D})$ } 
  \hspace*{\algorithmicindent} \textbf{Input:} Set of datasets $\mathbf{D} = \{\train_0, ..., \train_M\}$ from the same causal system under different contexts
  \\
  \hspace*{\algorithmicindent} \textbf{Input:} Kernel $k$, schedules for $\alpha_t, \beta_t$, and stepsizes $\eta_t$
  \\
  \hspace*{\algorithmicindent} \textbf{Output:} Set of CBN and intervention particles $\{(\G^{(l)}, \param^{(l)}, \I^{\text{tar}, (l)}, \param^{(l)}_{\I}) \}_{l=1}^L$
  \vspace*{0pt}
    \begin{algorithmic}[1] 
     % \begin{algorithmic}[lines] `lines` controls the line numbering: 0 means no line numbering, 1 means number every line, and n means number lines n, 2n, 3n.
    \State Initialize set of latent and parameter particles $\{\mathbf{\Omega}^{(l)}_0\}_{l=1}^L = \{ ( \Z^{(l)}_0, \intvGam^{(l)}_0, \param^{(l)}_0, \param^{(l)}_{\I, 0} )\}_{l=1}^L$ 
    \For{iteration $t = 0$ to $T - 1$}
    	\State Estimate $\nabla \log p(\Z, \param, \intvGam, \param_{\I} | \mathbf{D})$ for each $\mathbf{\Omega}^{(l)}_t = (\Z^{(l)}_t, \param^{(l)}_t, \intvGam^{(l)}_t, \param^{(l)}_{\I, t})$ \Comment{see Eq \ref{eq:score_Z}, \ref{eq:score_interv_gamma}}
    	    	
    	\For{particle $m = l$ to $L$}

        \State $\displaystyle \Z^{(l)}_{t+1} \leftarrow \Z^{(l)}_{t} +\eta_t ~ \phib^{\Z}_t(\mathbf{\Omega}^{(l)}_t)$ \Comment{SVGD steps}
        \Statex \quad \quad \quad \quad   where $\displaystyle \phib^{\Z}_t(\cdot) := \frac{1}{L} \sum_{l=L}^M 
			\Big [  
			 k \left(\mathbf{\Omega}^{(l)}_t, ~\cdot~ \right)  ~ \nabla_{\Z^{(l)}_t} \log p(\mathbf{\Omega}^{(l)}_t | \mathbf{D})
			+ \nabla_{\Z^{(l)}_t} k \left(\mathbf{\Omega}^{(l)}_t, ~\cdot~ \right)
			\Big ]$
		\State $\displaystyle \param^{(l)}_{t+1} \leftarrow \param^{(l)}_{t} +\eta_t ~ \phib^{\param}_t(\mathbf{\Omega}^{(l)}_t)$
        \State $\displaystyle \intvGam^{(l)}_{t+1} \leftarrow \intvGam^{(l)}_{t} +\eta_t ~ \phib^{\intvGam}_t(\mathbf{\Omega}^{(l)}_t)$
       	\State $\displaystyle \param^{(l)}_{\I, t+1} \leftarrow \param^{(l)}_{\I, t} +\eta_t ~ \phib^{\param_\I}_t(\mathbf{\Omega}^{(l)}_t)$ 
       	\Statex \quad \quad \quad \quad where $\phib_t^{\param}, \phib_t^{\intvGam}, \phib_t^{\param_\I}$ are analogous to $\phib_t^{\Z}$ but use gradients $\nabla_{\param^{(l)}_t}, \nabla_{\intvGam^{(l)}_t}, \nabla_{\param^{(l)}_{\I, t+1}}$
        \EndFor
      
    \EndFor 
    \State \Return $\{ (\G_\infty (\Z^{(l)}_T), \param^{(l)}_T, \I^{\text{tar}}_\infty (\intvGam^{(l)}_{T}),  \param^{(l)}_{\I, T})  \}_{l=1}^L$ \Comment{see Appx. \ref{appx:annealing}}
    \end{algorithmic}
    \label{alg:bacadi_svgd}
\end{algorithm}

\section{MARGINAL INFERENCE WITH THE BGE SCORE}
\label{appx:bge_theory}
In addition to joint posterior inference that includes the {\em parameters} $\param$ of the model, we also consider the {\em marginal} posterior $p(\G|\mathbf{D})$. We employ the commonly used Bayesian Gaussian Equivalent (BGe) marginal likelihood that scores Markov equivalent structures equally \citep{geiger1994learning, geiger2002parameter}. This model factorises the marginal likelihood into components for each node given its parents. Details on the computation of the BGe score are provided by \citet{kuipers2014addendum}. Other good explanations are given by \citet{grzegorczyk2010introduction} and \citet{kuipers2022iBGE}.

Compared to the case of considering parameters $\param$, the marginal likelihood does not yield factorization over different datasets. That is,
\begin{align*}
p(\mathbf{D} | \G, \I^{\text{tar}})  \neq \prod_{k=1}^{M} p(\mathcal{D}_k|\G, \Itar_k)
\end{align*}
Instead, we have to consider the \emph{fused} data from all contexts by
\begin{align}
    \mathbf{X} &= \begin{bmatrix}
           \rvx_{1}^T \\
           \rvx_{2}^T \\
           \vdots \\
           \rvx_{N}^T
         \end{bmatrix} \in \mathbb{R}^{N\times d} \text{ \hspace{.2cm} with \hspace{.1cm} }
    \mathbf{c} = \begin{bmatrix}
           c_{1} \\
           c_{2} \\
           \vdots \\
           c_{N}
         \end{bmatrix}  \in \{0,\dots,M\}^{N}
\end{align}
for $N=\sum_{k=1}^M n_k$ and variables $c_i \in \{0, \dots, M\}$ that indicate from which context sample $\rvx_i$ originates. As before, $c_i=0$ denotes the observational context. 

\textbf{Interventions. } When considering hard interventions, we cut off the connections of a variable to its parents. This effectively means that the data at this variable only contributes a constant (wrt. the current hypothesis graph) factor to the likelihood, and thus the scoring of a hypothesis graph should not be affected. In other words, when computing the score for a certain variable $j$ in the graph $\G$, we drop all datapoints in $\mathbf{X}$ where $j$ was the target of a hard intervention and then compute the BGe score for the remaining datapoints $\hat{\mathbf{X}}_j$. In our implementation, we achieve this by masking out datapoints in $\mathbf{X}$. This enables us to still make use of the Gumbel-Softmax estimator for the Bernoulli intervention targets. 

\textbf{Priors. } Following the notation of \citet{geiger2002parameter} and \citet{kuipers2014addendum}, we use the standard effective sample size hyperparameters $\alpha_\mu=1$ and $\alpha_\omega=d+2$. Moreover, we use the diagonal form as the Wishart inverse scale matrix for the Normal-Wishart parameter prior underlying the BGe score (cf. \citep{grzegorczyk2010introduction}) with a prior mean of ${{\bm{\mu}}}=[0,\dots,0]^T$.

For interventions, we use the same approach but set the prior values $\bm{\mu}$ to the real intervention mean from which the interventions where sampled. We then compute a BGe score for the intervened datapoints given an empty graph in order to have comparable likelihood. This helps when estimating the intervention targets.

\begin{figure}[t]
    \centering
    \includegraphics[width=.7\linewidth]{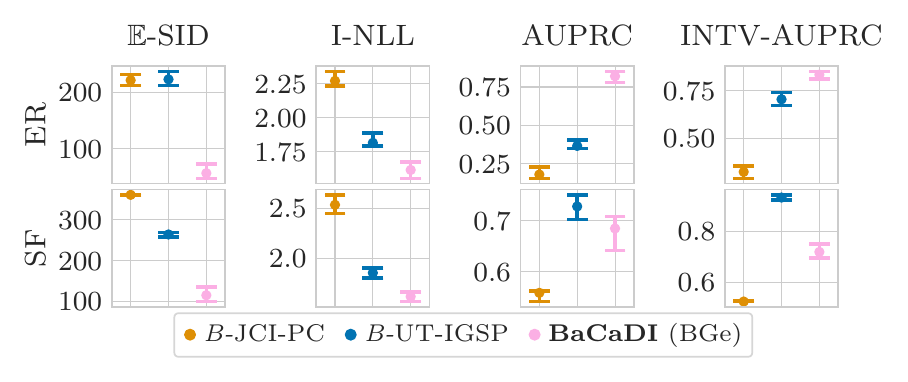}
    \caption{Additional results for Linear Gaussian BNs with $d=20$ nodes. Here, we use the BGe score \citep{geiger1994learning, geiger2002parameter} that marginalizes out the parameters of a linear Gaussian model. We again see how \ours  better predicts causal mechanisms across all settings. Note that DCDI-G is not included since it always performs joint inference with parameters.}
    \label{fig:bge}
\end{figure}

\textbf{Experiments. } We evaluate \ours  using the marginal BGe score to the baselines of JCI-PC and UT-IGSP. Note that the algorithms of both JCI-PC and UT-IGSP are not affected by the different score; the difference is the scoring of the predicted graph structures, where we compute a likelihood score using the marginal BGe (replacing the MLE parameter estimate). We do not compare to DCDI-G as it always performs joint inference with parameters. Similar to Sec. \ref{sec:eval}, we focus on linear Gaussian BNs with $d=20$ nodes. We report the results in Fig. \ref{fig:bge}. As done previously, we select the models with the lowest heldout I-NLL.

\label{appx:experiments_bge}

We again see how \ours  outperforms related methods in most metrics, particularly $\mathbb{E}$-SID and I-NLL. Only UT-IGSP performs competitively for AUPRC and INTV-AUPRC of SF graph structures, but fails to be on par with \ours  in other settings.
These results resonate with the other experimental evaluations, showing promising results for making causal structure predictions.

%%%%%%%%%%%%%%%%%%%%%%%%%%%%%%%%%%%%%%%%%%%%%%%%%%%%%%%%%%%%%%
\section{EXPERIMENTAL SETUP}
\label{appx:exp_details}
In the following, we describe the models, datasets and implementations that were used to perform experiments in detail. The code is available under \url{https://github.com/haeggee/bacadi}.

\subsection{Model Details \& \ours }
\label{appx:model_details}
First, we discuss the models of graphs as well as the exact usage of \ours \footnote{Some parts of the model descriptions follow the appendix in \citep{lorch2021dibs} and are included for completeness.}.

\textbf{Graphs. }
We consider DAGs that follow either \erdosrenyi~ (ER) \citep{gilbert1959random} or scale-free (SF) \citep{barabasi1999emergence} distributions. ER graphs follow a prior distribution 
\begin{equation}
\label{eq:er_prior}
p(\G) \propto q^{\|\G\|_{1}}(1-q)^{{{d}\choose{2}} -\|\G\|_{1}}
\end{equation}
that describes that each edge exists independently w.p. $q$. For SF graphs, we define a prior
\begin{equation}
\label{eq:sf_prior}
p(\G) \propto \prod_{i=1}^{d}(1+\|\G_i\|_{1})^{-3}
\end{equation}
analogous to their power law degree distribution $p(\text{deg}) \sim \text{deg}^{-3}$. $\G_i$ describes the $i$-th row of the adjacency matrix. For all our experiments, we use priors that result in $2d$ edges in expectation. Building on \citet{lorch2021dibs}, \ours can incorporate such a graph distribution in the prior $p(\Z)$ (see Sec. 4.2 in their paper).

\textbf{Overview: Gaussian BNs. } As instantiations of \ours's inference models that describe the local conditional distributions, we consider Bayesian networks with additive Gaussian noise. This means that the variables $\rvx=(\evx_1,\dots,\evx_d)$ follow a distribution 
\begin{equation}
    \label{appx_eq:gaussian_bn}
    p(\evx_i|\param, \G) = \mathcal{N}(f(\evx_{\pa{i}}, \param), \sigma_i)
\end{equation}
For the inference with \ours , we assume a fixed observation noise $\sigma_i^2=\sigma^2=0.1$.

The function $f$ can be modelled in different ways as follows.

\textbf{Linear BNs. } Linear Gaussian BNs model the mean of a given variable as a linear function of its parents:
\begin{equation}
    \label{appx_eq:lingauss_bn}
    p(\rvx|\param, \G) = \mathcal{N}((\G \circ \param)^T\rvx, \sigma \mathbf{I}),
\end{equation}
where $\circ$ denotes the element-wise multiplication. We use this parametrization as it allows for constant dimensionality of $\param \in \mathbb{R}^{d\times d}$ and the elementwise multiplication only keeps the parents of each variable. Moreover, this allows the use of the Gumbel-Softmax estimator in Eq. \ref{eq:score_Z}.

As a prior, we use a standard Gaussian $p(\param_{i,j}|\G_{i,j}=1)=\mathcal{N}(0,1)$.

\textbf{Nonlinear BNs. } The local conditionals can be extended to \emph{nonlinear} dependencies modelled by neural networks. We consider feed-forward neural networks of the form 
\begin{equation}
    \text{FFN}(\rvx;\param) := \param^{(L)}\sigmoid(\dots,\param^{(1)}\rvx+{\theta}_b^{(1)}) \dots)+{\theta}_b^{(L)},
\end{equation}
where $\param=((\param^{(L)}, \theta_b^{(L)}), \dots, (\param^{(1)}, \theta_b^{(1)}))$ describe the parameters of the $l$-th layer for $l\in \{1,\dots,L\}$. The function $\sigmoid$ is the element-wise nonlinear activation function and $\theta_b^{(l)}$ is the bias. This then gives the distribution
\begin{equation}
    \label{appx_eq:fcgauss_bn}
    p(\rvx|\param, \G) = \prod_{i=1}^{d} \mathcal{N}(\evx_i; \text{FFN}(\G_i^T \circ \rvx; \param_i), \sigma).
\end{equation}
Note that we thus have one neural network defined by $\param_i$ for each of the local conditional distributions of variable $i$, that is, $d$ networks in total. For all experiments, we use one hidden layer with $5$ units and the sigmoid activation function.

As a prior for the parameters, we analogously use a standard Gaussian with mean 0 and variance 1.

\textbf{Interventions. } We model interventions using a Gaussian likelihood $p_i^{I_k}(\evx_i| \param_{I_k}) = \mathcal{N}(x_i|\mu^I_{k,i}, \sigma_I)$ with fixed variance ${\sigma_I}^2=0.5$. We infer the means $\param_{I_k} = [\mu^I_{k,1}, ..., \mu^I_{k,d}]$ for which we set a wide Gaussian prior $p(\param_{I_k} | \Itar_k) = \prod_{i \in \Itar_k} p(\mu^I_{k,i}| I^{\text{tar}}_{k,i}=1) = \prod_{i \in \Itar_k} \mathcal{N}(\mu^I_{k,i}|0, 10)$. This reflects an uninformative prior over a large effect range of the interventions.

\textbf{Initialization. } For the linear Gaussians, the parameters are initialized closed to zero via $\param_{\text{init}} \sim \mathcal{N}(0, \sigmoid_{\text{init}}\mathbf{I})$ with $\sigmoid_{\text{init}}=0.3$. The closeness to zero is important to avoid inducing a bias at the start of the SVGD inference process. Similarly, we sample $\param_{\I,\text{init}} \sim \mathcal{N}(0, \sigmoid_{\I,\text{init}}\mathbf{I})$ with $\sigmoid_{\I, \text{init}}^2=0.1$. 

For nonlinear Gaussians, we use the Glorot (sometimes called Xavier) normal to initialize the weights of the neural networks. \citep{glorot2010understanding}.

\textbf{SVGD. } We instantiate \ours with the SVGD algorithm as discussed in Sec. \ref{sec:svgd_instantiation} and shown in Alg. \ref{alg:bacadi_svgd}. In all our experiments, we use 20 particles, which matches the number of bootstrap samples for the baselines (see \ref{appx:baselines}).

\subsection{Data Generation for the Synthetic Causal Inference Tasks}
\label{appx:data_generation}
The data generation is done as follows: we first sample a random graph (either ER or SF) and then sample random parameters. We collect data samples by iterating through the topological ordering of the graph and sample a variable given its local parents. Since we consider additive Gaussian models, each variable is described by a distribution of the form $\evx_i | \param, \G \sim \mathcal{N}(f(\evx_{\pa{i}}, \param),\sigma_i)$ where $f$ is either a linear function or a nonlinear feedforward neural network (cf. Sec. \ref{appx:model_details}). The noise variables $\sigma_i$ are sampled per variable and fixed once from $\sigma_i^{2}\sim\mathcal{U}[0.05, 0.15]$. If the noise variables had the same variance across all variables in the graph, this would render identification possible \citep{peters2014identifiability}. 

\vspaceparagraph
\textbf{Parameters.} The parameters and generative models are initialized as follows:
\begin{itemize}[wide=0pt, leftmargin=8pt]
    \item \emph{Linear BNs:} We sample the parameters $\param$ uniformly and independently from $\mathcal{U}([-2,-0.5]\cup[0.5,2])$ in order to bound the weights sufficiently away from zero.
    \item \emph{Nonlinear BNs:}
    The NNs for each local conditional are the same model as used for \ours  as well as DCDI-G, that is, a fully connected NN with single hidden layer of size of $5$ with biases. The nonlinear activations are sigmoid functions. All weights and biases are drawn randomly and independently from a Gaussian $\mathcal{N}(0,1)$.
\end{itemize}

\textbf{Interventions. }
As described in the main text, we perform {\em hard} interventions on every node for the $20$-node graphs. We create random values by first sampling $\hat{\mu}_{k,i}\sim\mathcal{N}(0,2)$ and then setting ${\mu}_{k,i}^I=\text{sign}(\hat{\mu}_{k,i}) \cdot 5 + \hat{\mu}_{k,i}$. This ensures that the interventions performed are bounded away from zero and out-of-distribution. If a variable $x_i$ is the target of the intervention in context $k$, we then have $p_i^{I_k}(\evx_i| \param_{I_k}) = p_i^{I_k}(\evx_i| {\mu}_{k,i}^I) = \mathcal{N}(x_i; \mu_{k,i}^I, \sigma_I)$, where $\sigma_I^2=0.5$.

\subsection{Data Generation with the SERGIO Gene-Expression Simulator}
\label{appx:sergio_details}
SERGIO \citep{dibaeinia2020SERGIO} is a single-cell expression simulator for gene regulatory networks (GRNs). The software tool can generate realistic single-cell transcriptomics datasets based on a user-defined graph input that describes the regulatory network.
SERGIO uses stochastic differential equations to simulate a gene's expression dynamics as a function of the changing levels of its regulators. The simulations resemble the data collected by modern high-throughput, single-cell RNA sequencing (scRNA-seq) technologies and are thus more practically-relevant than simulators approximating ``bulk'' microarray data \citep{schaffter2011genenetweaver}. 
For more details, we refer to the original paper by \citet{dibaeinia2020SERGIO}. In the following, we give a brief overview of how we simulate scRNA-seq data with SERGIO.
For this, we use an implementation that is adapted from the open-source code available under {\texttt{https://github.com/PayamDiba/SERGIO}}, which is published under a GPL-3.0 license. 

\textbf{Simulation. } SERGIO generates synthetic scRNA-seq data $\mathbf{D}$ for a given causal graph with $d$ genes by first simulating clean gene expression data and then corrupting these expressions with technical measurement noise.
The $N$ observations in $\mathbf{D}$ correspond to $N$ cell samples, i.e. one row in $\mathbf{D}$ describes the joint expression of the $d$ genes in a single cell.
For simplicity, we do not add the technical noise modules in our experiments.

To simulate the single-cell gene expressions, SERGIO samples randomly-timed expression snapshots from the steady state of the stochastic dynamical system modeling the GRN. In this regulatory process, the genes are expressed at rates influenced by other genes using the chemical Langevin equation. The source nodes in the causal graph $\G$ are denoted master regulators (MRs), whose expressions evolve at constant production and decay rates. The downstream gene expressions evolve non-linearly under production rates caused by the expression of their causal parents in $\G$. 
To model different cell types, SERGIO varies the specifications of the MR production rates, which significantly influence the evolution of the system. 
As a result, the data contains variation across cell types and due to biological noise within cells of the same type. In our experiments, we generate single-cell samples collected from ten cell types \citep{dibaeinia2020SERGIO}.

% In the second stage, the clean gene expressions sampled previously are corrupted with technical noise that resembles the noise phenomena found in real scRNA-seq data. For simplicity, we do not make use of the technical noise for our experiments.

\textbf{Interventions. } We consider \emph{knockout interventions} that clip the expression of a specific gene to zero. To this end, we extend SERGIO by forcing the production rate of knocked-out genes to be zero during simulation. As described in the main text, we do this for $M=10$ gene targets and arrive at $11$ contexts, including the observational setting. 

\textbf{Parameters. }
To specify the simulation parameters of SERGIO, we follow the settings suggested by \citet{dibaeinia2020SERGIO} for synthetic data generation.
Given a causal graph $\G$, the simulation for $c$ cell types of $d$ genes is governed by the following parameters:
\begin{itemize}[wide=0pt, leftmargin=8pt]
\item $k \in \mathbb{R}^{d \times d}, k_{i,j} \sim \mathcal{U}[1,5]$ : interaction strength if edge $i \rightarrow j$ exists in $\G$

\item $b \in \mathbb{R}_{+}^{d \times c}, b_{i,j}\sim \mathcal{U}[1,3]:$ MR production rates if $j$ is a source node in $\G$ 

\item $\gamma \in \mathbb{R}_{+}^{d \times d}, \gamma_{i,j}=2$ : Hill function coefficients controlling nonlinearity of interactions

\item $\lambda \in \mathbb{R}^{d}, \lambda_i=0.8$ : decay rates per gene

\item $\zeta \in \mathbb{R}_{+}^{d}, \zeta_i=1.0$ : scale of stochastic process noise in chemical Langevin equation

\end{itemize}

% We set these parameters

We standardize the data for all methods by subtracting the empirical mean and dividing by the standard deviation.

% The data generation is influenced by the following parameters:
% \TODO{fix this list of items and make it formal?}
% \begin{itemize}[wide=0pt, leftmargin=8pt]
% \item $k \in \mathbb{R}^{d\times d}$ is signed float matrix parameterising the strength of the causal dependencies between the gene expressions
% \item hills is [d,d] matrix, whose entries parameterise the nonlinear Hill function regulating each causal effect (in practice all entries are the same)
% \item b is a [d, c] matrix, where c is the number of cell types we simulate.
% Each of the c columns parameterises the basal reproduction rates for each of the genes for one cell type, which determines the expression of a gene if it is a source node. (If not a source node, the simulator ignores the entry of b)
% Thus, in this simulator, a given cell type is defined solely through a row in b, the rest (i.e. k or hills) is the same. In the simulator, c is also called “bins” sometimes.
% \item The total number of samples we sample will be N = c * M, where M is the number of single cells simulated per cell type adThus there are 2 sources of variations/noise across samples: biological noise of the SDEs within a cell type, and variation across cell types. In general, only different cell types really provide signal. but since the paper only considers up to c <= 9, I would suggest we fix c=10 and M=10 or 20 or so.
% \end{itemize}

\subsection{Baselines \& Implementation}
\label{appx:baselines}
In this section, we provide additional details on the baseline methods and their implementations used in our experiments.
% Our source-code and instructions on how to reproduce results are available under \url{https://www.dropbox.com/sh/5vtj4zp9h8sr9zq/AACkJ5naAxhXnpIByNEezI4Ia?dl=0}.

For all methods, we use implementations adapted from the source code of DCDI available under \url{https://github.com/slachapelle/dcdi} (where the authors also included their baselines' code). The implementation of JCI-PC is a modified version of the R package using code from the JCI repository \url{https://github.com/caus-am/jci/tree/master/jci}. UT-IGSP has an implementation available from \url{https://github.com/uhlerlab/causaldag}.

\textbf{DAG Bootstrap. } To compare all methods in Bayesian model averaging, we employ the nonparametric DAG bootstrap \citep{friedman2013data}. It performs model averaging by bootstrapping the observations $\mathbf{D}$ to yield a collection of synthetic data sets, each of which is used to learn a single graph. We sample with replacement from $\mathbf{D}$ for each dataset $\mathcal{D}_k$ individually. The collection of unique single graphs approximates the posterior by weighting each graph by its unnormalized posterior probability.

All of the methods are evaluated with 20 bootstrap samples, i.e. the same number of samples as particles used for the SVGD instantiation of \ours. For DCDI-G, we only use 5 bootstrap samples due to the significantly longer runtimes.

\textbf{JCI-PC. }
The Joint Causal Inference (JCI) framework \citep{mooij2016multiple} introduces a general formulation to extend the initial causal graph by auxiliary nodes that describe the different contexts, effectively performing causal discovery over a graph of size $d+M$. This approach can be instiantiated with different standard algorithms. In our experiments, we use the PC algorithm \citep{spirtes2000causation}, which relies on conditional independence tests (CI) to discover the Markov Equivalence Class (MEC), i.e. the skeleton of a graph with v-structures as well as possible identifiable edge directions. The PC algorithm uses a Gaussian CI test with significance threshold $\alpha^{\text{JCI}}$, which are best-suited for Gaussian BNs. 

For computing the graph metrics, we compute the $\mathbb{E}$-SID between the CPDAGs of the GT graph and predictions of JCI-PC. Additionally, we favor JCI-PC when computing AUPRC scores. See Sec. \ref{appx:experiment_metrics} for more details.

To arrive at a DAG that we can use to estimate MLE parameters and compute log-likelihood metrics, we generate a random consistent expansion of the CPDAG as defined by \citet{chickering2002optimal} using the algorithm by \citet{dor1992simple}. That is, we generate a DAG s.t. the CPDAG has the same skeleton and v-structures and every directed edge in the CPDAG has the same direction in the DAG. 

\textbf{UT-IGSP. }
The interventional greedy sparsest permutation (IGSP) method \citep{wang2017permutation, yang2018characterizing} proposes an algorithm that learns causal structures via local scores based on CI relations and permutation search. The work is extended by \citet{squires2020permutationbased} to the case of unknown targets (UT-IGSP). Analogous to JCI-PC, IGSP uses Gaussian CI and invariance tests with parameters $\alpha^{\text{UT-IGSP}}$ and $\alpha^{\text{UT-IGSP}}_{\text{inv}}$. Since we study the low sample setting, it is possible that the algorithm as provided by their open source implementation fails to compute an essential correlation matrix. Specifically, the algorithm computes a correlation matrix that becomes singular in the case when the number of unique samples is smaller than the number of variables considered, thus rendering an inversion of this matrix impossible. In case this happens, we retry the inference with a halved $\alpha^{\text{UT-IGSP}}$ confidence threshold. The maximum number of restarts is set to $10$, and the bootstrap sample is dropped in case this maximum is reached. 

Similar to JCI-PC, the $\mathbb{E}$-SID is computed as the midpoint of the lower and upper bound between the CPDAGs of the GT graph and predictions of UT-IGSP. As for JCI-PC, we favor UT-IGSP when computing AUPRC scores. See Sec. \ref{appx:experiment_metrics} for more details.
Likewise, we obtain a DAG by generating a random consistent expansion of the CPDAG to compute the log-likelihood metrics.

\textbf{DCDI-G. }
\citet{brouillard2020differentiable} introduced Differentiable Causal Discovery from Interventional Data (DCDI), which performs causal structure learning via the augmented Lagrangian method. The algorithm formulates a continuous-constrained optimization problem that relies on stochastic gradient descent and neural networks to fit the local conditionals.

To ensure a fair comparison, we employ the exact same likelihood model as used for the nonlinear Gaussian BNs used in \ours, that is, a feedforward neural network with one hidden layer of size $5$ and Gaussian additive noise. This model is called DCDI-G in \citep{brouillard2020differentiable}.
However, we use a smaller neural network to model the means of the local conditional distributions than originally used by \citet{brouillard2020differentiable}. In initial experiments with DCDI-G, we observed strong overfitting due to our low-sample regime and thus reduced the model capacity. 
As suggested by \citet{brouillard2020differentiable}, we use the leaky ReLU with negative slope of $0.25$ as elementwise activation function. To obtain comparable log-likelihood metrics, we fix the noise variables to $\sigma^2=0.1$ as done for \ours. These noise variables could generally be learned by both methods.

% \looseness-1 Since DCDI-G takes longer to compute, we restrict the number of bootstrap samples to 5.

\subsection{Metrics} \label{appx:experiment_metrics}
Our reported metrics focus on three essential aspects of our inference problem: causal graph prediction, intervention detection, and learning the local conditional distributions of the individual variables.

\begin{itemize}[wide=0pt, leftmargin=8pt]
\item {\bf SID:} The {\em Structural Interventional Distance} (SID) \citep{peters2015structural} quantifies the closeness between two DAGs in terms of how well their interventional adjustment sets coincide. Since we perform posterior inference and arrive at a distribution over graphs, we consider the \emph{expected} SID, which is defined as

\begin{equation}
\mathbb{E}\text{-SID}(p, \G_{\text{gt}}) := \sum_{\G} p(\G|\D) \cdot \text{SID}(\G, \G_{\text{gt}})
\end{equation}

To compute the SID, we use the implementation provided by the Causal Discovery Toolbox \citep{kalainathan2019cdt}.
Since UT-IGSP and JCI-PC only return a CPDAG of the Interventional Markov Equivalence Class (I-MEC), we calculate the lower and upper bound of the SIDs in the I-MEC and report their midpoint as the $\mathbb{E}$-SID. The DAG bootstrap variants for all baselines as well as \ours  use the weighted mixture rather than the empirical distribution of samples. The weight is based on the unnormalized log-likelihood achieved on the bootstrap sample.

\item {\bf SHD}: The structural hamming distance (SHD) reflects the graph edit distance to the ground truth DAG. However, it often does not properly reflect the closeness of two DAGs in terms of their causal interpretation. For instance, the trivial prediction of the empty graph achieves competitive SHD scores for the sparse graphs we consider in this paper. In our main text, we thus focus on the $\mathbb{E}$-SID as well as other metrics that together better assess the quality of causal graph predictions. For completeness, we report the $\mathbb{E}$-SHD results in Appx. \ref{appx:eshd}.

\item {\bf Threshold metrics}: Treating the edge prediction as a classification task, we compute the area under the precision recall curve (AUPRC) for pairwise edge prediction based on the posterior marginals $p(g_{ij}=1| \mathbf{D})$. This marginal is defined as the posterior mean of an indicator variable for the presence of the edge: $p(g_{ij}=1| \mathbf{D}) = \mathbb{E}_{p(\G|\mathbf{D})} \mathbf{1}[g_{ij}=1]$. 
The AUPRC provides more insight into predictive performance than, e.g., the AUROC because sparse graphs translate to highly imbalanced edge classification tasks.

For JCI-PC and UT-IGSP, which both return a CPDAG with edges that are undecided, we favor the methods when computing the AUPRC metric. 
Specifically, we orient a predicted undirected edge correctly when a ground truth edge exists and only count a falsely predicted undirected edge as a single mistake.

\item {\bf Interventional AUPRC}: Similarly, we report the {\em interventional} AUPRC (INTV-AUPRC) for the classification of intervention targets. The INTV-AUPRC captures how well an algorithm predicts which variables have been intervened on. Since we assume sparse interventions, this metric better captures a model's performance for the imbalanced classification task.

\item {\bf I-NLL:} We compute the average negative {\em interventional log-likelihood (I-LL)} on $M^{\text{test}} = 10$ heldout interventional datasets $\mathbf{D}^{\text{test}} = \{\train^{\text{test}}_1, ..., \train^{\text{test}}_{10} \}$, where different interventions are performed compared to the training datasets. Each interventional test dataset comprises 100 samples and has known intervention targets $\Itar_{\text{test}, k}$ and effect distributions $p(x_i|\param_{I_{\text{test}, k}})$

The I-NLL is computed via
\begin{align*}
    \text{I-NLL}(p, \mathbf{D}^{\text{test}}) := -  \frac{1}{M^{\text{test}}} \sum_{k=1}^{M^{\text{test}}}  \E_{p(\G, \param | \mathbf{D})} \left[ \frac{1}{|\train^{\text{test}}_k|} \log p(\train^{\text{test}}_k | \G,\param,\Itar_{\text{test}, k}, \param_{I_{\text{test}, k}}) \right] \;.
\end{align*}
Since UT-IGSP and JCI-PC are not equipped with local conditional distributions, we use the linear Gaussian maximum-likelihood parameters (MLE), which can be computed in closed-form \citep{Hauser_2014}. For the nonlinear datasets, this creates a model mismatch since the closed form is only available for linear Gaussian BNs.
\end{itemize}

\subsection{Hyperparameter Search}
To ensure a fair comparison, we perform a hyperparameter search for all baselines. The search ranges can be found in Table \ref{tab:hyperparameter_searchspace}. Throughout all experiments, we use at least 20 hyperparameter samples and aggregate results over 30 different random seeds and graphs. For the results of \ours in the main text, we fix the hyperparameters $\alpha=0.01$, $\beta=2$ and $\lambda=1$, which are the linear scale for the temperature parameter of the sigmoids, the linear scale for the sparsity regularizer and the sparse intervention regularizer, respectively. 

\begin{table}[ht!]
    \centering
    \begin{tabular}{c | c | c | c}
        \toprule
        {Method}                 & {Hyperparameter}                       & Comment                                & Search range                      \\
        \midrule
        \multirow{3}{*}{\ours~}   %& $\alpha$                               & Linear slope temperature for Bernoulli        & $\log_{10} \mathcal{U}[-3, -1]$              \\
                                %  & $\beta$                                & Linear slope temperature for acyclicity prior &   $\mathcal{U}[1, 3]$   \\
                                 & $\tau_Z$                               & Kernel lengthscale                     & $\log_{10} \mathcal{U}[-1, 1.7]$  \\
                                 & $\tau_\gamma$                          & Kernel lengthscale                     & $\log_{10} \mathcal{U}[-1, 1.7]$  \\
                                 & $\tau_\theta$                          & Kernel lengthscale                     & $\log_{10} \mathcal{U}[1.2, 5]$   \\
                                %  & $\lambda$                              & Intervention sparsity prior            & $\log_{10} \mathcal{U}[-1, 2]$    \\

        \\[-1.1em]
        \midrule
        \\[-1.1em]

        \multirow{1}{*}{JCI-PC}  & $\alpha^{\text{JCI}}$                  & CI tests                               & $\log_{10} \mathcal{U}[-5, -1]$   \\
        \\[-1.1em]
        \midrule
        \\[-1.1em]
        \multirow{2}{*}{UT-IGSP} & $\alpha^{\text{UT-IGSP}}$              & CI tests                               & $\log_{10} \mathcal{U}[-5, -1]$   \\
                                 & $\alpha^{\text{UT-IGSP}}_{\text{inv}}$ & Invariance tests                       & $\log_{10} \mathcal{U}[-5, -1]$   \\
        \\[-1.1em]
        \midrule
        \\[-1.1em]
        \multirow{4}{*}{DCDI-G}  & batch size                             & -                                      & $\mathcal{U}(\{16,32,64\})$ \\
                                 & $\lambda_R$                            & Sparsity coefficient for interventions & $\log_{10} \mathcal{U}[-8, -1]$   \\
                                 & $\lambda$                              & Sparsity coefficient for graph         & $\log_{10} \mathcal{U}[-3, -1]$   \\
                                 & $h$                                      & Convergence threshold                  & $\log_{10} \mathcal{U}[-8, -6]$   \\
        \bottomrule
    \end{tabular}
    \vspace{5pt}
    \caption{Hyperparameter search space for all methods}
    \label{tab:hyperparameter_searchspace}
\end{table}

% \newpage
\section{ADDITIONAL EXPERIMENTS}
\label{appx:additional_experiments}

\subsection{Larger datasets}

\begin{figure}[h!]
    \centering
    \vspacefigabove
    \includegraphics[width=.7\linewidth]{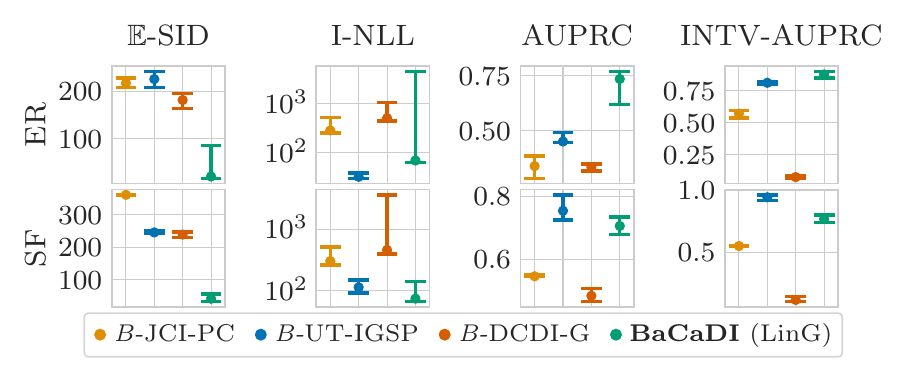}
    \caption{{\bf{Linear Gaussian with more data.}} Joint posterior inference over CBNs and interventions for {\em linear Gaussian} ground-truth CBNs. The results are for data from ER-2 (top) and SF-2 (bottom) graphs with $d=20$ variables and $M = 20$ contexts. Here, we double the dataset size to $N=600$ samples in total. Similar to the previous setting with lower sample sizes, \ours  is consistently the most competitive method.
    %For each metric, we report the median and the 90\% confidence interval.
    \vspacefigcaptionlow
    }
    \label{fig:lingauss_moredata}
\end{figure}

\begin{figure}[h!]
    \centering
    \vspacefigabove
    \includegraphics[width=.7\linewidth]{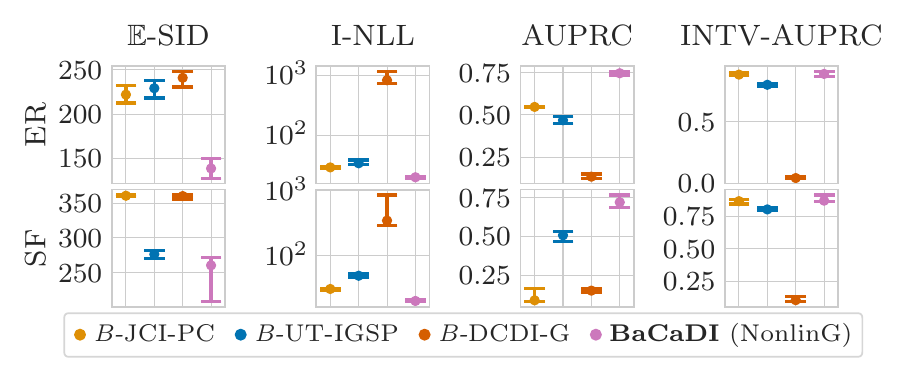}
    \caption{{\bf{Nonlinear Gaussian with more data.}} Joint posterior inference over CBNs and interventions for {\em nonlinear Gaussian} ground-truth CBNs. The results are for data from ER-2 (top) and SF-2 (bottom) graphs with $d=20$ variables and $M = 20$ contexts. Here, we double the dataset size to $N=600$ samples in total.  \ours  performs the best across all metrics. 
    %For each metric, we report the median and the 90\% confidence interval.
    \vspacefigcaptionlow
    }
    \label{fig:fcgauss_moredata}
\end{figure}

We report additional results when doubling the dataset size, i.e. we collect $n_0=200$ observational samples and $n_k=20$ samples per interventional context. The results for synthetic linear and nonlinear Gaussian BNs can be found in Fig. \ref{fig:lingauss_moredata} and Fig. \ref{fig:fcgauss_moredata}, respectively.

\subsection{Observational vs. Interventional Data}
\label{appx_sec:obs_vs_intv}
As an additional interesting use case, we evaluate how interventional data helps predicting the causal structure with \ours. In general, intervening on variables in the system and observing the outcome provides information that helps discovering the causal mechanisms.
In the perfect setting, when the intervention targets are fully known, they increase identifiability by shrinking the interventional Markov equivalence class \citep{hauser2012characterization}. However, the information gain can be limited when the intervention targets are unknown \citep{squires2020permutationbased}. 

\looseness-1 We investigate how \ours  can leverage such unknown interventions compared to using just observational data. To that end, we evaluate 3 different settings: i) when only observational data is available, ii) interventional data with full knowledge of the interventions, and iii) unknown interventions. When the interventions are known, the posterior inference of  \ours  can be reduced to Eq. \ref{eq:posterior_known_interventions}.

\begin{figure}[t]
    \centering
    \includegraphics[width=.9\linewidth]{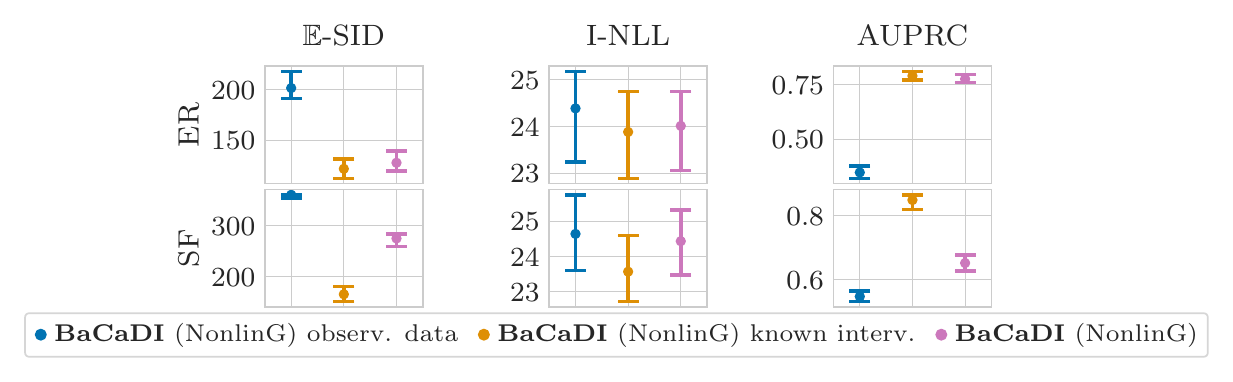}
    \caption{{\bf{Comparison: Observational data vs. known and unknown interventions.}} Additional results comparing \ours  with observational data against known and unknown interventions for nonlinear Gaussian BNs for $d=20$ node graphs. As a natural baseline, the setting of knowing the intervention targets and effects leads to the best performance across all metrics. When the interventions are unknown, \ours  achieves results close to this baseline. Notably, it outperforms the setting where just observational data is available.}
    \label{fig:obs_vs_interv}
\end{figure}

\looseness-1 We consider nonlinear Gaussian BNs for $d=20$ node graphs. All methods receive the exact same number of samples for inference. That is, we collect $n_0=300$ samples from the ground-truth BN without interventions for the observational dataset.
As before, the interventional datasets have $n_0=100$ observations and $n_k=10$ samples per interventional context. We report the results in Fig. \ref{fig:obs_vs_interv}.

We observe that \ours  achieves significantly better results when interventional data is available. The case of known interventions serves as a natural baseline, where \ours  achieves the best performance and gets closest in recovering the true causal structure. This is most clearly shown by the $\mathbb{E}$-SID and AUPRC scores. Notably, \ours  is able to get close this baseline even when interventions are unknown and outperforms the setting where just observational data is available. This shows the benefit of performing and collecting interventions even when some of the effects may be unknown and is a promising result for future work.

\subsection{50 Node graphs}
\begin{figure}[t]
    \centering
    \includegraphics[width=.7\linewidth]{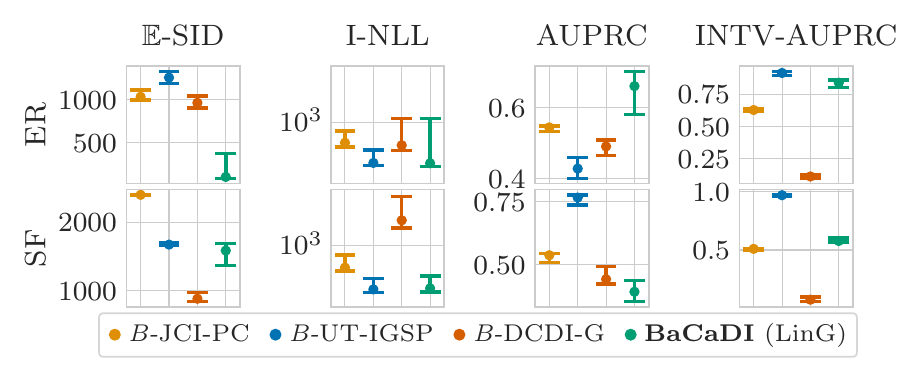}
    \caption{{\bf{Linear Gaussian BNs with 50 nodes.}} Additional results comparing \ours  on larger graphs with $d=50$ nodes for Linear Gaussian BNs for ER-2 (top) and SF-2 (bottom). \ours  performs competitively across all metrics. In particular, it makes significantly better causal mechanism predictions for ER graph as captured by the $\mathbb{E}$-SID.}
    \label{fig:lingauss_50node}
\end{figure}
As additional experiments, we perform posterior inference for larger graphs with $50$ nodes in total. Analogous to the previous evaluations, we consider synthetic linear Gaussian BNs with hard interventions on every node. We use a larger dataset of $n_0=200$ observational samples and $n_k=20$ samples per interventional context. The results are given in Fig.~\ref{fig:lingauss_50node}. We find that \ours  scales to larger graphs and performs competitively across all metrics. In particular, our method outperforms the baselines by a large margin in terms of the $\mathbb{E}$-SID and AUPRC for ER graphs.

\subsection{Structural Hamming Distance}
\label{appx:eshd}
For completeness, we include the {\em Expected Structural Hamming Distance} ($\mathbb{E}$-SHD) metric for all results discussed in the main text for $d=20$ node graphs in Table \ref{tab:eshd}. Analogous to the $\mathbb{E}$-SID, the $\mathbb{E}$-SHD is defined as $\mathbb{E}\text{-SHD}(p, \G_{\text{gt}}) := \sum_{\G} p(\G|\D) \cdot \text{SHD}(\G, \G_{\text{gt}})$. The SHD reflects the graph edit distance where wrongly inverted edges are counted as only one error. However, while it captures closeness to the ground truth graph, trivial predictions like the empty graph achieve competitive results for sparse graphs.
For example, in the setting of $d=20$ nodes and $2d$ edges in expectation, the empty graph will achieve a $\mathbb{E}$-SHD of $40$ in expectation. Similar conclusions hold if only very few edges are predicted. In the main text, we thus resort to the $\mathbb{E}$-SID as well as additional metrics that together report an accurate description of the quality of inference.
\begin{table}[ht!]
    \centering
    \begin{tabular}{l  c | c | c | c | c}
        \toprule
        \\[-1.1em]
                   & JCI-PC         & UT-IGSP        & DCDI-G           & \ours~ (LinG)   & \ours~ (NonlinG)  \\
        \\[-1.1em]
        \toprule
        \\[-1.1em]
        \multicolumn{6}{l}{\bf{Linear Gaussian BNs graphs:}}                                                       \\
        \midrule
        \\[-1.1em]
        ER-2              & 38.97 {\small ($ 1.65$) } & 41.38 {\small ($ 2.53$) } & 63.29 {\small ($ 4.52$) } & 50.32 {\small ($10.98$) } & - \\
        SF-2            & 37.42 {\small ($ 0.32$) } & 34.35 {\small ($ 1.18$) } & 51.68 {\small ($ 3.67$) } & 36.48 {\small ($ 6.12$) } & - \\
        \\[-1.1em]
        \midrule
        \\[-1.1em]
        \multicolumn{6}{l}{\bf{Nonlinear Gaussian BNs graphs:}}                                                    \\
        \\[-1.1em]
        \midrule
        \\[-1.1em]
        ER-2              & 38.93 {\small ($ 1.65$) } & 35.09 {\small ($ 1.76$) } & 64.91 {\small ($ 2.61$) } & - & 18.73 {\small ($ 1.85$) } \\
        SF-2            & 37.00 {\small ($ 0.00$) } & 33.75 {\small ($ 0.48$) } & 57.92 {\small ($ 2.69$) } & - & 23.55 {\small ($ 1.43$) } \\
        \\[-1.1em]
        \midrule
        \\[-1.1em]
        \multicolumn{6}{l}{\bf{SERGIO graphs}}                                                                     \\
        \\[-1.1em]
        \midrule
        \\[-1.1em]
        SF-2            & 44.29 {\small ($ 0.70$) } & 45.05 {\small ($ 1.88$) } & 56.39 {\small ($ 1.63$) } & 108.71 {\small ($ 3.75$) }        & 115.04 {\small ($ 6.22$) }        \\
        \bottomrule
    \end{tabular}
    \vspace{10pt}
    \caption{\textbf{Expected Structural Hamming Distance. } We report the $\mathbb{E}$-SHD for all methods for the main results with $d=20$ node graphs. We aggregate results over 30 different seeds and report the mean and standard error. While the $\mathbb{E}$-SHD is a simple and commonly used metric, it can be misleading. For sparse graphs, the trivial prediction of the empty graph achieves competitive $\mathbb{E}$-SHD scores; similar assessments can be made for predictions with only a few edges. We include the results for completeness.}
    \label{tab:eshd}
\end{table}

\subsection{Runtimes}
\label{appx:run_times}
We report the average runtimes (in minutes) for all methods and main results for $d=20$ node graphs in Table \ref{tab:runtimes}.
\begin{table}[h!]
    \centering
    \begin{tabular}{l  c | c | c | c | c}
        \toprule
        \\[-1.1em]
                   & JCI-PC         & UT-IGSP        & DCDI-G           & \ours~ (LinG)   & \ours~ (NonlinG)  \\
        \\[-1.1em]
        \toprule
        \\[-1.1em]
        \multicolumn{6}{l}{\bf{Linear Gaussian BNs graphs:}}                                                       \\
        \midrule
        \\[-1.1em]
        {ER}       & 2.07$\pm$ 0.16 & 1.31$\pm$ 0.22 & 260.03$\pm$56.11 & 51.68$\pm$13.12 & -                 \\
        SF-2       & 2.17$\pm$ 0.15 & 1.46$\pm$ 0.09 & 176.89$\pm$30.46 & 60.57$\pm$22.95 & -                 \\
        \\[-1.1em]
        \midrule
        \\[-1.1em]
        \multicolumn{6}{l}{\bf{Nonlinear Gaussian BNs graphs:}}                                                    \\
        \\[-1.1em]
        \midrule
        \\[-1.1em]
        ER         & 1.89$\pm$ 0.07 & 1.16$\pm$ 0.06 & 143.88$\pm$12.53 & -               & 491.26$\pm$91.74  \\
        SF-2       & 1.92$\pm$ 0.06 & 1.41$\pm$ 0.06 & 145.43$\pm$25.83 & -               & 519.79$\pm$97.95  \\
        \\[-1.1em]
        \midrule
        \\[-1.1em]
        \multicolumn{6}{l}{\bf{SERGIO graphs}}                                                                     \\
        \\[-1.1em]
        \midrule
        \\[-1.1em]
        SF-2       & 2.36$\pm$ 0.09 & 0.42$\pm$ 0.20 & 128.04$\pm$ 9.97 & 45.11$\pm$ 1.52 & 323.11$\pm$101.15 \\
        \bottomrule
    \end{tabular}
    \vspace{10pt}
    \caption{\textbf{Average runtimes} for all methods for the main results with $d=20$ node graphs. We aggregate results over 30 different seeds and report the mean and standard deviation. All numbers given are in minutes.}
    \label{tab:runtimes}
\end{table}

\subsection{Computational Resources}
\label{appx:exp_compute}
All experiments reported in this paper were performed in bulk and in parallel on 2-core CPU nodes of an internal cluster. No GPUs were used. We estimate the total compute to amount to roughly 40,000 hours of CPU time.
\newpage
\section{STEIN VARIATIONAL GRADIENT DESCENT}
\label{appx:svgd}
We here give an overview of Stein Variational Gradient Descent (SVGD) introduced by \citet{liu2016stein}. Fundamentally, the work of \citet{liu2016stein} connects the mathematical notions of probability discrepancies with a variational inference method, which closely resembles the gradient descent algorithm. For details, we refer to the original paper.

\paragraph{Stein's identity and discrepancy} Formally, let $p(\rvx)$ be a continuously differentiable (i.e. smooth) density supported on $\mathcal{X}\subseteq \mathbb{R}^d$. For a smooth vector function $\phi(\rvx)$, \emph{Stein's identity} states that for sufficiently regular $\phi$, we have
\begin{align}
    \mathbb{E}_{\rvx\sim p}[\mathcal{A}_p \phi(\rvx)] = 0, \text{ where } & \mathcal{A}_p \phi(\rvx)= \phi(\rvx)\nabla_x \log p(\rvx)^{T} + \nabla_x \phi(\rvx)
\end{align}
where $\mathcal{A}_p$ is the Stein operator acting on the function $\phi$. This equation can be subsequently used as a discrepancy measure: when considering the expectation over a smooth density $q$ different than $p$, we obtain the so called \emph{Stein discrepancy} by considering the \emph{maximum violation of Stein's identity}. That is, we have 
\begin{align}
    \mathbb{S}(q,p)=\max_{\phi \in \mathcal{F}} [\mathbb{E}_{\rvx\sim q}[\text{trace}(\mathcal{A}_p \phi(\rvx))]^2]
\end{align}
for a choice of a set of functions $\mathcal{F}$, for instance the reproducing kernel Hilbert space (RKHS) denoted by $\mathcal{H}^d$.

\paragraph{Variational Inference using smooth transforms} The goal of variational inference is to approximate a target distribution $p$ using a simpler distribution $q^*$ which minimizes the KL-divergence
\begin{align}
    q^* = \argmin_{q\in \mathcal{Q}} KL(q||p)
\end{align}
In order to minimize the KL-divergence, the authors in \citep{liu2016stein} consider incremental transforms formed by a small perturbation to the identity map that make up the set of distributions $\mathcal{Q}$. That is, the transform $\mathbf{T}: \mathcal{X} \rightarrow \mathcal{X}$ is defined as
\begin{align}
    \mathbf{T}(\rvx)=\rvx+\epsilon \phi(\rvx)
\end{align}
where $\phi$ is a smooth function that characterizes the perturbation direction. As a key result, \citet{liu2016stein} connect these transforms to the Stein operator and the derivative of the KL divergence. The authors show that if the function $\phi$ lies in the ball of the vector valued RKHS $\mathcal{H}^d$, the direction of the steepest descent on the KL divergence between a fixed $q$ and the target $p$ is given by
\begin{align}
    \phi_{q,p}^* = \mathbb{E}_{\rvx\sim q}[k(\rvx,\cdot) \nabla_{\rvx} \log p(\rvx) + \nabla_{\rvx} k(\rvx,\cdot)]
\end{align}
where one can easily identify the form of the Stein operator. What is more, the value of the obtained gradient equals the (negative) kernelized Stein discrepancy measure $-\mathbb{S}(q,p)$.

\paragraph{General Algorithm} This mathematical result suggests an iterative method that transforms an initial reference distribution $q_0$ to the target distribution $p$. Starting with a finite set of random particles $\{\rvx^{(m)}\}_{m=1}^{M}$, for some iteration $t$ each particle is updated deterministically according to
\begin{align}
    & \rvx_{t+1}^{(m)} \leftarrow \rvx_{t}^{(m)} + \epsilon_t \phi(\rvx_{t+1}^{(m)}) \\
    & \text{ where } \phi(\rvx) = \frac{1}{M} \sum_{l=1}^{M} k(\rvx_{t+1}^{(l)}, \rvx) \nabla_{\rvx} \log p(\rvx) + \nabla_{\rvx} k(\rvx_{t+1}^{(l)}, \rvx)
\end{align}
These steps iteratively decrease the KL divergence between $q_t$ and $p$, ultimately converging. 

Importantly, the advantage of SVGD is that it only depends on the gradient of the kernel $k(x,\cdot)$ that can be defined for the application, as well as the score function $\nabla_{\rvx}\log p(x)$ which can be computed without knowing the (intractable) normalization constant of $p$. On an intuitive level, the first part of the perturbation direction expression drives the particles to high density regions close to the mode of the target distribution $p$, whereas the term $\nabla_{\rvx} k(\rvx,\cdot)$ acts as a repulsive force that prevents the particles from collapsing together.

\end{document}